%% file: main.tex
\documentclass[11pt, letterpaper, logo, onecolumn, numbering]{gr}
\usepackage{etoolbox}

\usepackage[authoryear, sort&compress, round]{natbib}

\usepackage[inkscapeformat=png]{svg}

\usepackage[most, breakable, skins]{tcolorbox}
\usepackage{academicons}

\tcbuselibrary{skins}
\usepackage{lipsum}
\usepackage{CJKutf8}
\usepackage{placeins}
\usepackage{tabularx}
\usepackage{afterpage}
\usepackage{booktabs}
\usepackage{subcaption}
\usepackage{makecell}
\usepackage{multirow}
\usepackage{multicol} 
\usepackage{array}
\usepackage{float}
\usepackage{listings, listings-rust}
\usepackage{fontawesome5}
\usepackage{amssymb,graphicx}
\usepackage[dvipsnames]{xcolor}
\usepackage{hyperref}
\usepackage{cleveref}
\usepackage{longtable}
\usepackage{graphicx}
\usepackage{pdflscape}
\usepackage{adjustbox}
\usepackage{tikz}
\usetikzlibrary{calc,positioning,chains,shapes,arrows,fit,decorations.pathmorphing,patterns,fadings,shadows,patterns.meta,arrows.meta}
\usepackage{wrapfig}
\usepackage{dialogue}
\usepackage{algorithm}
\usepackage{algorithmic}
\usepackage{colortbl}
\usepackage{mdframed}

\usepackage{listings} 
\usepackage{CJKutf8}
\usepackage{tcolorbox} 
\usepackage[dvipsnames]{xcolor}
\usepackage{multicol}  
\usepackage{threeparttable}

\input{math_commands.tex}

\input{showcase_format}

\theoremstyle{plain}

\theoremstyle{definition}

\theoremstyle{remark}

\usepackage{CJKutf8}

\definecolor{medgray55}{gray}{0.55}
\definecolor{medgray}{gray}{0.7}
\definecolor{litegray}{gray}{0.9}
\definecolor{gblue}{RGB}{210, 227, 252}
\definecolor{gred}{RGB}{250, 210, 207}
\definecolor{gyellow}{RGB}{254, 239, 195}
\definecolor{ggreen}{RGB}{206, 234, 214}
\definecolor{gorange}{RGB}{254, 223, 200}

\definecolor{gblue9}{RGB}{23, 78, 166}
\definecolor{gred9}{RGB}{165, 14, 14}
\definecolor{gyellow9}{RGB}{227, 116, 0}
\definecolor{ggreen9}{RGB}{13, 101, 45}
\definecolor{gorange9}{RGB}{176, 96, 0}

\definecolor{myblue}{rgb}{0,0,1}
\definecolor{myred}{rgb}{1,0,0}
\definecolor{mylightgray}{gray}{0.95}

\definecolor{highlightblue}{HTML}{185ABC}

\usepackage{tcolorbox}
\tcbuselibrary{listings,skins,breakable}

\definecolor{codebg}{RGB}{252,252,253}
\definecolor{codeframe}{RGB}{205,209,217}
\definecolor{codetitle}{RGB}{232,236,244}
\definecolor{codetitletext}{RGB}{36,41,47}
\definecolor{codekw}{RGB}{170,13,145}
\definecolor{codestr}{RGB}{196,26,22}
\definecolor{codecom}{RGB}{106,115,125}
\definecolor{codenum}{RGB}{28,0,207}
\definecolor{codefn}{RGB}{38,71,124}
\definecolor{codebuiltin}{RGB}{0,112,32}

\lstdefinestyle{pythonclean}{%
  language=Python,
  basicstyle=\ttfamily\scriptsize,
  keywordstyle=\color{codekw}\bfseries,
  stringstyle=\color{codestr},
  commentstyle=\color{codecom}\itshape,
  identifierstyle=\color{black},
  emph={[2]def,return,if,else,for,in,not,and,or},
  emphstyle={[2]\color{codekw}\bfseries},
  emph={[3]np,array,reshape,self},
  emphstyle={[3]\color{codebuiltin}},
  emph={[4]kelly_criterion},
  emphstyle={[4]\color{codefn}\bfseries},
  frame=none,
  xleftmargin=0pt, xrightmargin=0pt,
  framexleftmargin=0pt, framexrightmargin=0pt,
  aboveskip=0pt, belowskip=0pt,
  showstringspaces=false,
  breaklines=true,
  columns=fullflexible,
  keepspaces=true,
  literate=%
    {0}{{{\color{codenum}0}}}1 {1}{{{\color{codenum}1}}}1
    {2}{{{\color{codenum}2}}}1 {3}{{{\color{codenum}3}}}1
    {4}{{{\color{codenum}4}}}1 {5}{{{\color{codenum}5}}}1
    {6}{{{\color{codenum}6}}}1 {7}{{{\color{codenum}7}}}1
    {8}{{{\color{codenum}8}}}1 {9}{{{\color{codenum}9}}}1
    {.0}{{{\color{codenum}.0}}}2 {.25}{{{\color{codenum}.25}}}3
    {£}{{\pounds}}1,
}

\newtcblisting{codepanel}[1]{%
  listing only, listing style=pythonclean,
  enhanced,
  arc=3pt, outer arc=3pt, boxrule=0.5pt,
  colback=codebg, colframe=codeframe,
  colbacktitle=codetitle, coltitle=codetitletext,
  fonttitle=\small\bfseries\sffamily,
  title={#1}, titlerule=0pt,
  toptitle=4pt, bottomtitle=4pt,
  top=5pt, bottom=5pt, left=6pt, right=6pt,
  boxsep=0pt,
  before skip=8pt, after skip=8pt,
  drop shadow={black!15},
}

\makeatletter

\renewcommand\paragraph{\@startsection{paragraph}{4}{\z@}%
            {-2.5ex\@plus -1ex \@minus -.25ex}%
            {1.25ex \@plus .25ex}%
            {\itshape\normalsize\bfseries}}
\makeatother
\newcolumntype{L}[1]{>{\raggedright\let\newline\\\arraybackslash\hspace{0pt}}m{#1}}
\newcolumntype{C}[1]{>{\centering}m{#1}}

\newcolumntype{R}[1]{>{\raggedleft\let\newline\\\arraybackslash\hspace{0pt}}m{#1}}

\definecolor{ao}{rgb}{0.0, 0.0, 1.0}

\newcommand\vcent[1]{\vcenter{\hbox{#1}}}

\newcommand\loudspeaker[1][3]{\ensuremath{\vcent{\rule{.6ex}{.6ex}}\kern-.5ex%
  \vcent{\scalebox{.6}[1]{\rotatebox[origin=center]{90}{$\blacktriangle$}}}%
  \ifnum#1>0\relax\kern.05ex\vcent{\scalebox{.4}{\ttfamily)}}%
  \ifnum#1>1\relax\kern-.4ex\vcent{\scalebox{.56}{\ttfamily)}}%
  \ifnum#1>2\relax\kern-.55ex\vcent{\scalebox{.7}{\ttfamily)}}%
  \fi\fi\fi}%
}

\definecolor{green}{rgb}{0.9,0.9,0.9}

\makeatletter
\renewcommand\subparagraph{%
 \@startsection {subparagraph}{5}{\z@ }{3.25ex \@plus 1ex
 \@minus .2ex}{-1em}{\normalfont \normalsize \bfseries }}%
\makeatother

\let\cite\citep

\title{KellyBench: A Benchmark for Long-Horizon Sequential Decision Making}

\reportnumber{} 

\renewcommand{\today}{}

\author[$1$]{Thomas Grady}
\author[$1$$ $]{Kip Parker}
\author[$1$$ $]{Iliyan Zarov}
\author[$1$$ $]{Henry Course}
\author[$1$$ $]{Chengxi Taylor}
\author[$1$$\dagger$]{Ross Taylor}
\affil[$1$]{General Reasoning, Inc.}
\affil[$\dagger$]{Correspondence to hello@gr.inc}

\begin{abstract}
Language models are saturating benchmarks for procedural tasks with narrow objectives. But they are increasingly being deployed in long-horizon, non-stationary environments with open-ended goals. In this paper we introduce \texttt{KellyBench}, an environment for evaluating sequential decision-making in sports betting markets. Agents are placed in a sequential simulation of the 2023–24 English Premier League season and tasked with maximising their long-term bankroll growth. They are given detailed historical data, including advanced statistics, lineups, and public odds. To succeed they must build machine learning models, identify edge in public markets, and adapt as the environment changes over time. We find that all frontier models evaluated lose money on average over the course of the season for five seeds. The best performing model achieves an average return of -8\%, and many models experiencing ruin across seeds. To judge strategy sophistication, we use a human expert rubric to grade each model and find their approaches to be unsophisticated compared to human baselines; Claude Opus 4.6 achieves a rubric score of 26.5\%, which means there is significant room for improvement.  \texttt{KellyBench} is available as an open-access API endpoint at \url{https://openreward.ai/GeneralReasoning/KellyBench}.
\end{abstract}

\begin{document}
 \maketitle

\vspace{2em}

\begin{figure}[H]
\centering
    \includegraphics[width=1.00\linewidth]{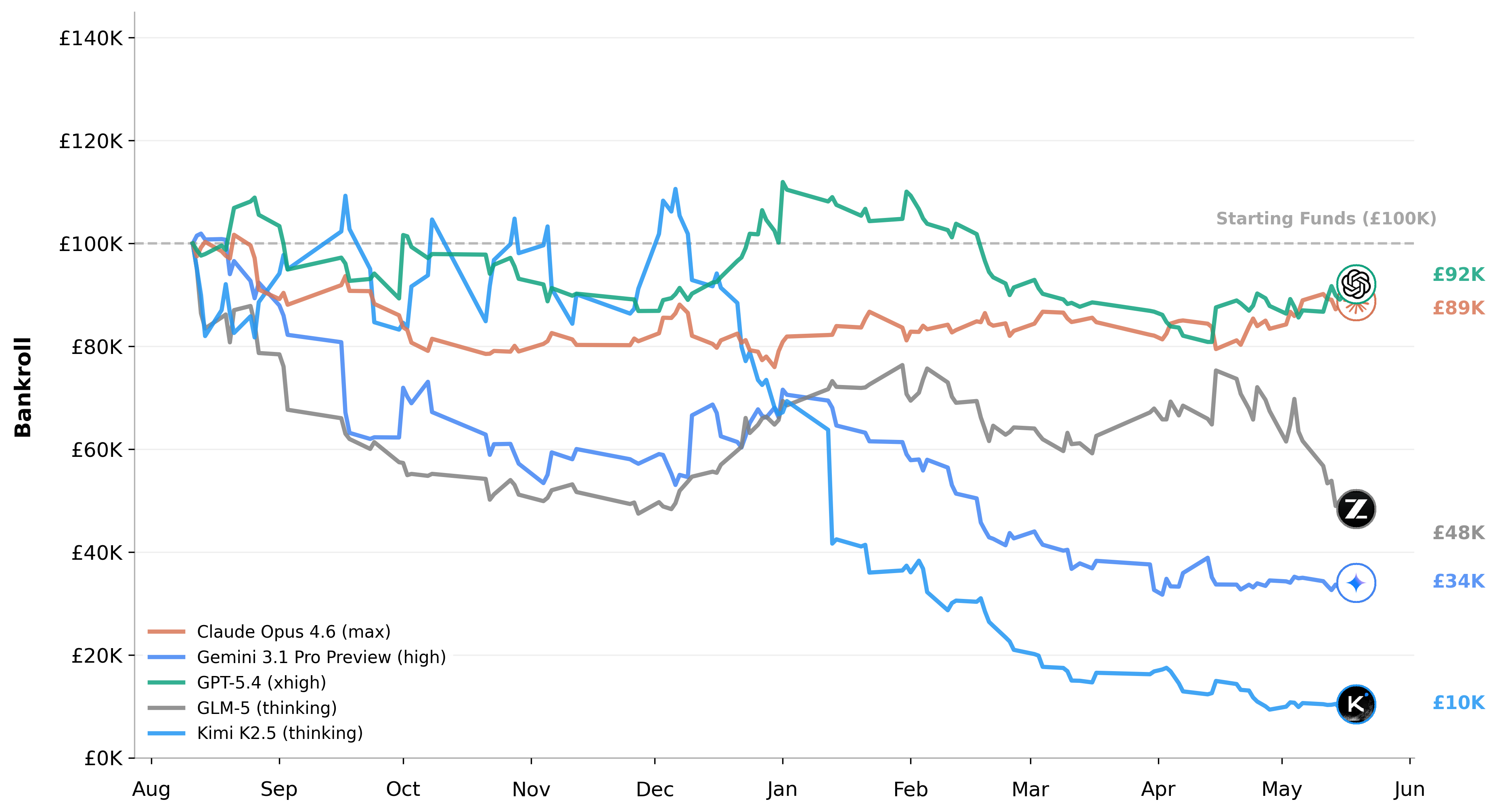}
    \caption{\textbf{Model Performance on KellyBench}. KellyBench tasks models with developing machine learning betting strategies for the 2023/24 English Premier League season with the goal of maximising long-term bankroll growth. No model makes a return on average across 5 seeds. Models also fail to adapt strategies in response to failure. Initial bankroll is normalised to £100K for display purposes.}
\label{fig:sigmoid_steepness_comparison}
\end{figure}

\begin{figure}[h]
\centering
    \includegraphics[width=1.00\linewidth]{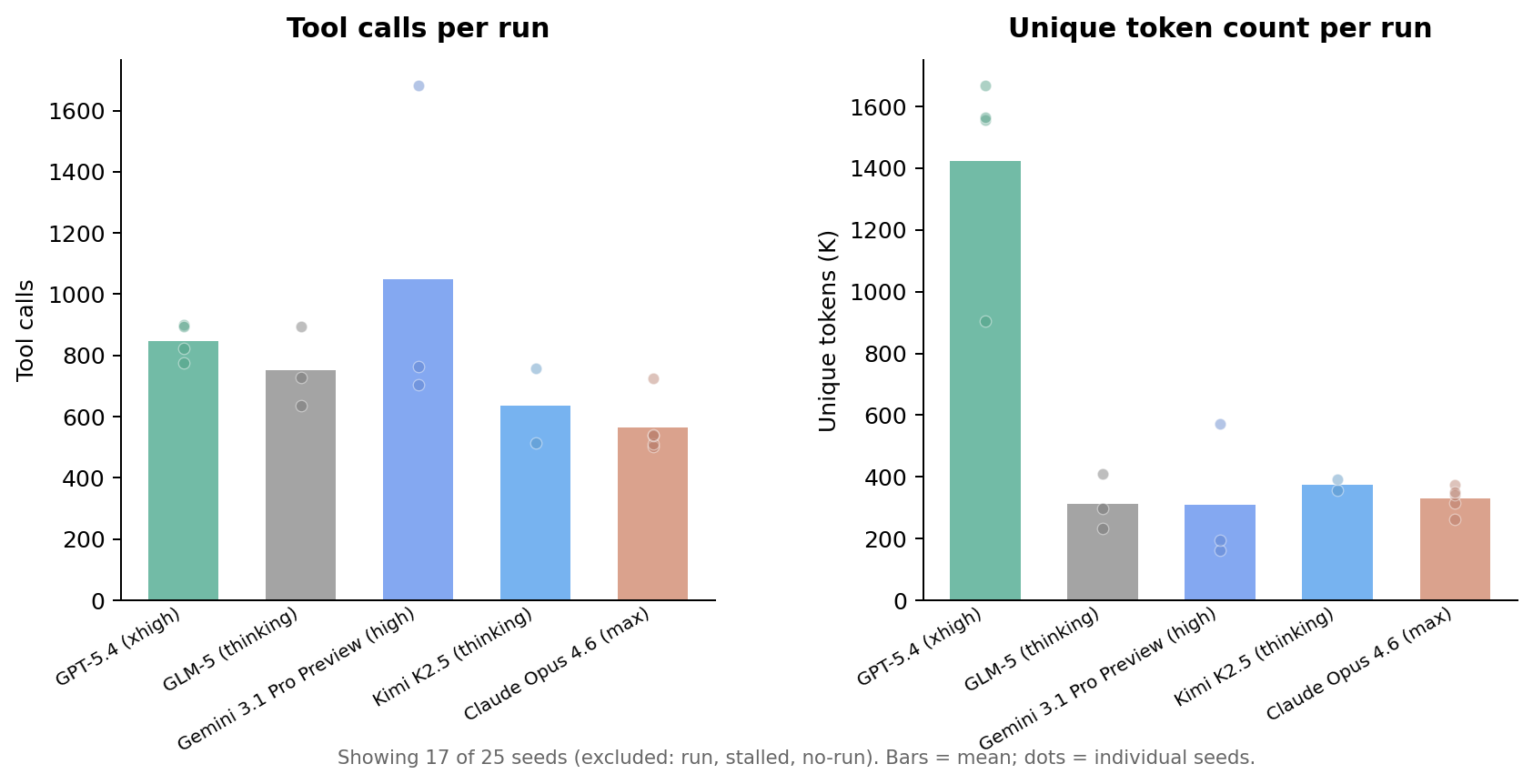}
    \caption{\textbf{Long-Horizon Sequential Decision Making}. Current frontier models we evaluated in \texttt{KellyBench} used 500-1000 tool calls to complete an episode and around 100K-1.7M unique tokens.}
\label{fig:sigmoid_steepness_comparison}
\end{figure}

\newpage

\section{Introduction}
\label{sec:intro}
\input{intro}

\section{Background}
\label{sec:background}
\input{background}

\section{Benchmark Design}
\label{sec:benchmark-design}
\input{benchmark-design}

\section{Results}
\label{sec:findings}
\input{findings}

\section{Limitations and Conclusions}
\label{sec:lim}
\input{limitations}

\section{Acknowledgements}

Thanks to Anthony Hartshorn, Ran Achiron and Nick Levine for their helpful comments on the initial draft of the paper.

\bibliography{sample}

\newpage 
\input{app}

\end{document}

%% file: math_commands.tex
\usepackage{amsmath,amsfonts,bm}

\def\eqref#1{equation~\ref{#1}}

\def\1{\bm{1}}

\DeclareMathAlphabet{\mathsfit}{\encodingdefault}{\sfdefault}{m}{sl}
\SetMathAlphabet{\mathsfit}{bold}{\encodingdefault}{\sfdefault}{bx}{n}



%% file: showcase_format.tex
\definecolor{ababcol}{HTML}{F14738}
\definecolor{myhailuo2}{HTML}{F97669}
\definecolor{querycol}{HTML}{7964E8}
\definecolor{goldanswercol}{HTML}{FFB43B}
\definecolor{otherscol}{HTML}{FC5BCF}
\definecolor{myhailuo3light}{HTML}{FFA9FA}
\input{colors}
\tcbset{
    showcase/.style={
        fonttitle=\large,
        colback=white!20,  
        colframe=black,   
        coltitle=white,   
        boxrule=0.5mm,    
        arc=2mm,          
        outer arc=2mm,    
        left=1mm,         
        right=1mm,        
        top=1mm,          
        bottom=1mm,       
        width=\textwidth, 
        before skip=0.1pt,
        after skip=0.1pt,
    },
    context/.style={
        fontupper=\scriptsize,
        fonttitle=\large,
        colframe=querycol,     
        coltitle=white,   
        colback=white,    
        boxrule=0.3mm,    
        arc=2mm,          
        outer arc=2mm,    
        left=1mm,         
        right=1mm,        
        top=1mm,          
        bottom=1mm,       
        before skip=1pt,
        after skip=0.1pt, 
    },
    query/.style={
        fontupper=\scriptsize,
        fontlower=\scriptsize,
        colframe=querycol,     
        coltitle=white,   
        colback=white,    
        boxrule=0.1mm,    
        arc=2mm,          
        outer arc=2mm,    
        left=1mm,         
        right=1mm,        
        top=1mm,          
        bottom=1mm,       
        before skip=1pt,
        after skip=0.1pt,
    },
    abab/.style={
        fontupper=\scriptsize,
        fonttitle=,
        colframe=ababcol, 
        coltitle=white,   
        boxrule=0.5mm,    
        arc=2mm,          
        outer arc=2mm,    
        left=1mm,         
        right=1mm,        
        top=1mm,          
        bottom=1mm,       
        width=0.33\textwidth, 
        before skip=0.1pt,
        after skip=0.1pt, 
    },
    others/.style={
        fontupper=\scriptsize,
        colframe=myhailuo3light, 
        coltitle=white,
        boxrule=0.5mm,    
        arc=2mm,          
        outer arc=2mm,    
        left=1mm,         
        right=1mm,        
        top=1mm,          
        bottom=1mm,       
        width=0.33\textwidth, 
        before skip=0.1pt,
        after skip=0.1pt, 
    },
    goldanswer/.style={
        fontupper=\scriptsize,
        colframe=goldanswercol,     
        coltitle=white,   
        boxrule=0.5mm,    
        arc=2mm,          
        outer arc=2mm,    
        left=1mm,         
        right=1mm,        
        top=1mm,          
        bottom=1mm,       
        width=0.33\textwidth, 
        before skip=0.1pt,
        after skip=0.1pt, 
    },
}

%% file: colors.tex
\definecolor{myhailuo1dark}{HTML}{FC8900}
\definecolor{myhailuo2dark}{HTML}{F14738}
\definecolor{myhailuo3dark}{HTML}{D12AAA}
\definecolor{myhailuo4dark}{HTML}{4C4DC2}

\definecolor{myhailuo1}{HTML}{FFB43B}
\definecolor{myhailuo2}{HTML}{F97669}
\definecolor{myhailuo3}{HTML}{FC5BCF}
\definecolor{myhailuo4}{HTML}{7964E8}

\definecolor{myhailuo1light}{HTML}{FFD085}
\definecolor{myhailuo2light}{HTML}{FFA19F}
\definecolor{myhailuo3light}{HTML}{FFA9FA}
\definecolor{myhailuo4light}{HTML}{BDACFB}

\colorlet{myorange}{Orange!20}
\colorlet{mygreen}{LimeGreen!25}
\colorlet{myyellow}{Yellow!30}
\colorlet{myblue}{CornflowerBlue!25}
\colorlet{mybrown}{RawSienna!25}
\colorlet{mypurple}{Orchid!25}
\colorlet{myred}{Red!60}
\colorlet{myorangefull}{YellowOrange!60}
\colorlet{mybrownfull}{RawSienna!60}

\colorlet{myorangethick}{Orange!40}
\colorlet{mygreenthick}{LimeGreen!50}
\colorlet{myyellowthick}{Yellow!60}
\colorlet{mybluethick}{CornflowerBlue!50}

%% file: intro.tex
One definition of intelligence is inductive inference, the ability to learn from experience, build models from acquired knowledge, and use those models for prediction \cite{hutter2007universalInductionIntelligence}. However, in the real world agents must convert predictions into useful actions. Betting markets provide an example of this difference. A predictive model that is more accurate than the public on average need not be profitable, and can lead to ruin if the edge on some bets is wrongly estimated \cite{benter2008computer, thorp1997kelly}.

\noindent Many popular evaluations of language models do not measure intelligence in this sense of learning from experience. Instead, they typically consider stationary environments, well-specified tasks, and sparse end-of-episode feedback. For example, one task in the popular \texttt{terminalbench2} evaluation asks an agent to "implement an adaptive-rejection sampler as described in Gilks et al. (1992)". While this tests procedural competence, it does not test the ability to formulate and revise models in light of experience \cite{merrill2026terminalbenchbenchmarkingagentshard,hughes2024openendednessessentialartificialsuperhuman}.

\noindent The real world is also non-stationary where the underlying rules change over time. However, most existing benchmarks have fixed behaviours. For example, the knight and the bishop have exactly the same behaviour in any game of chess, but a financial security will change behaviour under a new market regime, and an athlete's ability will change after a long-term injury \cite{AngTimmermann2012Regime, Johns2021Achilles}. In European football, for example, home advantage was found to decline in strength at the end of the 2019-2020 season under pandemic crowd restrictions, creating a bias in older predictive models \cite{Hill2021HomeFieldAdvantage}. This suggests a potential capability gap between models acting in static environments at training-time and dynamic environments at test-time.

\noindent To study these issues, we introduce \texttt{KellyBench}. \texttt{KellyBench} is an open-ended, non-stationary environment for measuring the ability of language models to make money in sports betting markets. \texttt{KellyBench} uses real market odds from the 2023/2024 English Premier League season and asks agents to bet from a bankroll for each matchday. Agents are given extensive historical data, including advanced statistics, lineups, and past market odds. They must develop machine learning models, identify edge relative to the market, and manage risk so as to maximise long-run bankroll growth.

\noindent Every model we evaluate on \texttt{KellyBench} loses money on average over the course of the season for five seeds, and several models experience ruin. The best performing model, GPT-5.4, achieves an average return on investment of $-8\%$. Only $3/25$ model seeds finish the simulation with a positive return, and all of these models have a negative return when averaged across 5 seeds. Via qualitative analysis of their trajectories, we find that models have poor adaptivity and low competence in accounting for potential estimation error and non-stationarity. In other words, the current generation of frontier models cannot consistently beat the market in our evaluation setup.

\noindent We also introduce a novel process-based measure of competence called \textit{sophistication}. Backtests can be subject to variance, so we consult experts with experience at quantitative betting funds to construct a 52-point rubric judging strategy sophistication. Using these rubrics, model strategies are consistently judged as unsophisticated relative to human baselines. The best performing model Claude Opus 4.6 has a sophistication score of $26.5\%$. Therefore, even with the limitations of our benchmark setup, and possibly high market efficiency, we believe there is considerable room for models to improve.

\noindent To examine whether our headline results are due to under-elicitation, we perform additional ablations with Claude Opus 4.6. We give Opus 4.6 access to relevant historical literature that human experts regard as seminal for constructing high-performing models, and also perform an evaluation using the Claude Code harness. In these ablations, neither intervention improves results and the model continues to lose money on average.

\noindent We conclude by discussing several limitations, including the possibility of underelicitation from the use of a single-agent harness, market efficiency, and the possibility that the data provided to the agent are insufficient. However, in spite of these limitations, our conclusion remains that models are systematically underperforming humans, and that much better performance is possible with the tools and data they are given. We argue a cultural shift is needed in making evaluations, moving away from fixed tasksets towards complex worlds where agents must learn from experience under uncertainty.

%% file: background.tex
\subsection{Quantitative Betting Approaches}

Probability theory arose in part from practical questions about gambling. For example, Cardano's sixteenth-century \textit{Liber de ludo aleae} combines advice on games of chance with reflections on probability and calculation \cite{bellhouse2005cardano}. Related ideas reappear in twentieth-century subjectivism, especially in Ramsey and de Finetti, where degrees of belief are interpreted through betting behaviour. In particular, Dutch book arguments are used to motivate Bayesian probability by excluding systems of belief that admit sure loss or "money pumps" \cite{ramsey1931truth, definetti1937prevision}.

\noindent The mathematical foundations of gambling were later reformulated in the mid-twentieth century through information theory. Shannon’s theory of communication provided a means to measure informational advantage \cite{shannon1948mathematical}. Kelly translated this into a repeated betting problem by showing that, when an agent has access to side information, the strategy that maximises expected log-wealth also maximises the long-run exponential growth rate of capital \cite{kelly1956new}. 

\noindent In particular, let \(X\) denote the outcome of a repeated gamble, \(Y\) the side information observed before betting, \(o(x)\) the gross odds on outcome \(x\), and \(b(x\mid y)\) the fraction of wealth staked on \(x\) after observing \(Y=y\). The agent seeks to maximise the expected logarithmic growth rate:

\[
W(b)=\sum_{x,y} p(x,y)\log\bigl(b(x\mid y)o(x)\bigr).
\]

\noindent In the canonical model, the optimal strategy is to bet in proportion to the conditional probabilities \(b^{*}(x\mid y)=p(x\mid y)\). Under fair odds this gives:

\[
W^{*}=H(X)-H(X\mid Y)=I(X;Y),
\]

\noindent so that the value of side information is exactly the mutual information between \(Y\) and \(X\). Informational advantage is thus converted directly into long-run capital growth.

\noindent The same point may be expressed in divergence form. If market odds imply a distribution \(q\), the true distribution is \(p^{*}\), and the agent sizes bets according to beliefs \(p\), then the expected log-growth is:

\[
g(p;p^{*},q)=\sum_x p^{*}(x)\log\frac{p(x)}{q(x)}
          =D_{\mathrm{KL}}(p^{*}\|q)-D_{\mathrm{KL}}(p^{*}\|p).
\]

\noindent Growth is positive when the agent's model is closer to the truth than the market-implied distribution. Superior calibration becomes an informational edge which can be converted into compounding gains.

\noindent In practice, the Kelly criterion makes this link operational. For a binary gamble with win probability \(p\) and net odds \(r\), it prescribes staking a fraction of wealth:

\[
f^{*}=\frac{rp-(1-p)}{r}.
\]

\noindent These principles were first operationalised at scale by Ed Thorp, who treated gambling and finance as structurally similar problems of edge detection, stake sizing, and risk control. In \textit{Beat the Market} and later at Princeton/Newport, this framework was applied to hedged mispricings in warrants, convertibles and equities \cite{thorp1967beatmarket}. The same capital-growth logic that governed bet sizing in blackjack and sports betting was used to allocate capital across trading opportunities in securities markets \cite{thorp2005perspective,thorp2008kelly}.

\noindent In betting markets specifically, Bill Benter and Alan Woods supplied a more direct large-scale operationalisation of Kelly-style staking. Benter's account of Hong Kong racing describes a computer-based handicapping system built from fundamental variables and a logit model, which was then refined by combining the model's probabilities with the public's implied probabilities in order to reduce bias and identify genuine overlays \cite{benter2008computer}. Wagering was treated as a separate optimisation problem. Benter employed Kelly staking, but with two important modifications: fractional Kelly to reduce sensitivity to estimation error, and limits arising from the bettor's own market impact.

\noindent These examples suggest what is required of highly capable AI systems in betting markets. The task is not merely to construct an accurate fundamental model of the world, but to identify \emph{relative} edge against the market, and to distinguish genuine opportunity from estimation error. In this sense, the task extends beyond inductive inference to sequential decision-making under uncertainty.

\subsection{Machine Learning Models for Football Betting}

Statistical models play a large role in feature design due to the presence of non-stationarity. Team and player strengths evolve over time, and this has historically been handled more naturally by approaches such as statistical state-space models rather than modern machine learning methods \cite{harvey1989forecasting}.

\noindent For football prediction, \citet{dixon1997modelling} introduce a dynamic team-ability model based on exponentially weighted Poisson likelihoods. \citet{glickman1998state} develop a Bayesian state-space model for NFL scores, while \citet{crowder2002dynamic} model attacking and defensive strengths as latent processes in a non-normal state-space framework. Similarly, \citet{koopman2015dynamic} propose a dynamic Poisson model with stochastically evolving intensities and show that its forecasts can support profitable betting strategies.

\noindent Other features may be obtained through machine learning methods for constructing advanced statistics. An important example in football is the expected goals (\(xG\)) model, which estimates shot-conversion probabilities from event or tracking data. \citet{anzer2021goal} show that a strong \(xG\) model can be built using extreme gradient boosting on synchronised positional and event data, while later work by \citet{mead2023expected} and \citet{bandara2024predicting} improves prediction by adding features about the sequence of actions preceding each shot.

\noindent These dynamic-ability and advanced-statistics features can in turn be combined with a broader collection of fundamental features, including head-to-head records, weather, travel distance, stadium altitude, and many other facets of the world \cite{watanabe2017weather, vandamme2019home, borghesi2007weather, nichols2014travel, coleman2017travel}. The football betting problem therefore has a large design space to hypothesise and test, and many sources of novelty through the real world.

\subsection{Related Work}

\noindent \textbf{Machine Learning Engineering Evaluations}. Existing work has mostly considered narrow evaluations of procedural ability in fitting and evaluating models. MLE-Bench evaluates models on 75 narrowly scoped offline Kaggle competitions \cite{chan2025mlebench}. MLGym constructs 13 open-ended AI research tasks spanning areas such as supervised learning, reinforcement learning, and game theory \cite{nathani2025mlgym}. More recently, PostTrainBench focuses on post-training language models, asking agents to improve a base language model on a target benchmark under bounded compute \cite{rank2026posttrainbench}. KellyBench differs from these benchmarks in that it is non-stationary, has a larger potential feature space, and challenges agents to operationalise models in a sequential decision-making environment.

\noindent \textbf{Forecasting Evaluations}. Recent work has produced a growing list of forecasting benchmarks for AI models. ForecastBench introduces a dynamic benchmark of unresolved future-event questions \cite{karger2025forecastbench}. FutureX provides a large live benchmark for  predicting the future with daily updates \cite{zeng2025futurex}. Prophet Arena evaluates models on continuously collected live forecasting tasks with accuracy, calibration and economic-value metrics \cite{yang2026prophetarena}. Bench to the Future offers a pastcasting environment for repeatable evaluation of forecasting agents \cite{wildman2025bench}. Lastly, OpenForecaster evaluates forecasting performance on a held-out OpenForesight test set and on FutureX, reporting accuracy and calibration-oriented gains while using an offline news corpus to reduce temporal leakage during evaluation \cite{chandak2025openforecaster}.

\noindent Existing forecasting benchmarks provide valuable tests of inductive inference, but they  evaluate agents on sparse, heterogeneous, and one-off event questions. Because sports betting has many structurally similar events, \texttt{KellyBench} provides more opportunities to test if a model can consistently translate estimated edges from a model into decisions without being undone by overbetting or estimation error, and arguably provides a better measure of sequential decision-making over long time horizons.

\noindent \textbf{Trading Evaluations}. There is also a growing literature of financial trading environments. PyMarketSim introduces a limit-order-book simulation for agents \cite{mascioli2024marketsim}. MarS introduces an order-level market simulation engine built on a generative foundation model \cite{li2024mars}. FinRL Contests provide environments spanning stock trading, order execution and cryptocurrency trading  \cite{walid2025finrlcontests}. StockBench evaluates LLM agents in multi-month stock-trading environments using return and risk metrics \cite{chen2025stockbench}. AI-Trader introduces a live benchmark across U.S. equities and cryptocurrencies \cite{fan2025aitrader}. Lastly, TraderBench focuses on adversarial cryptocurrency and options markets \cite{yuan2026traderbench}.

\noindent A realistic trading environment would benefit from broad, timestamped access to news and market context. However, constructing a faithful offline approximation of the information set at any time period is difficult and risks either temporal leakage or overly narrowing the task. \texttt{KellyBench} occupies a useful middle ground. It supports repeated market decisions over a large class of structurally similar events, while allowing rich pre-match information - such as lineups, match and player statistics - to be provided offline in a timestamped form. This preserves open-ended exploration without requiring a full offline reconstruction of the internet.

%% file: benchmark-design.tex
We develop the \texttt{KellyBench} environment using the Open Reward Standard (ORS), a protocol for defining RL environments with tasks, tools and state \cite{openrewardstandard2026}. We serve the environment on OpenReward and use the \texttt{firehorse} library for running a ReSum harness on the models of interest \cite{wu2025resumunlockinglonghorizonsearch}\footnote{\url{https://github.com/GeneralReasoning/firehorse}}.

\subsection{Environment Dynamics}
\label{sec:dynamics}

Each episode simulates one full English Premier League season, spanning approximately 100-150 unique matchdays. The agent is initialised with a bankroll and proceeds through the season one matchday at a time. On each matchday~$t$, the interaction follows a fixed cycle:

\begin{enumerate}
    \item \textbf{Observation.} The agent views the set of matches $\mathcal{M}_t$ scheduled for that day, along with closing decimal odds (pre-match odds near kickoff) sourced from real bookmakers.
    \item \textbf{Model development.} The agent reads and writes files in a sandboxed compute environment to build predictive models and wagering strategies.
    \item \textbf{Bet placement.} For each match $m \in \mathcal{M}_t$, the agent places wagers on any of five types of bet: home win, draw, away win, over 2.5 total goals, or under 2.5 total goals\footnote{It would be more realistic to use Asian handicap markets for quantitative funds, but we use 1X2 and Over/Under markets given public data availability and simplicity of interpretation. Note there is also "vig" in the odds that we use, which is around 5.3\%, which makes the task significantly harder. We also use bookmakers odds rather than exchange odds as they have better public data availability. In a future version of the benchmark, we are intending to use exchange odds instead.}. Each bet specifies a stake drawn from the agent's current bankroll.
    \item \textbf{Settlement.} All bets are resolved against the actual match outcomes. Winning bets return the stake multiplied by the quoted decimal odds; losing bets forfeit the stake. The agent's bankroll is updated accordingly.
    \item \textbf{Data update.} After settlement, the agent receives the latest match results and detailed player-level statistics for the completed matchday, which it can incorporate into subsequent model iterations.
\end{enumerate}

\noindent The agent is required to place at least one bet per matchday to induce it to expend effort on model development rather than abstaining from the task. Penny bets are still allowed under this rule as a capital conservation strategy.

\subsection{Data}
\label{sec:data}

Agents are provided with two categories of historical data to develop machine learning models: match-level data and player-level data.

\subsubsection{Match-level data}

A longitudinal dataset of English Premier League matches spanning from the 1993--94 season to the start of the evaluation season. Each record contains the match date, home and away teams, and the full-time result (home/draw/away) with scoreline. Coverage broadens over time: half-time scores are available from 1995--96; shot counts, fouls, corners, cards, and the match referee from 2000--01; and pre-kickoff decimal odds from multiple bookmakers (including over/under and Asian handicap markets) from 2002-03 onwards.

\subsubsection{Player-level data}

We provide the model with a detailed dataset of per-match player statistics drawn from major European leagues and cup competitions (including the Premier League, Championship, La Liga, Serie~A, Bundesliga, Ligue~1, domestic cup competitions and UEFA club competitions) from 2008 onwards. Each record is linked to a specific fixture and contains team lineups together with individual player statistics such as goals, assists, minutes played, shots, cards, tackles, interceptions and expected goals (xG) where available; as well as player information such as their age, height and playing position.

\noindent Data disclosure of both data categories is \emph{progressive} during each episode. At the start of the season the agent has access only to historical seasons, and after each matchday the environment provides the latest results and player statistics for the matches just completed. This mirrors the real-world information structure faced by a quantitative fund.

\subsection{Tools}
\label{sec:tools}

The agent interacts with \texttt{KellyBench} through two types of tool: environment tools and CLI tools.

\subsubsection{Environment tools}

The agent interacts with the simulated betting world through four tools:

\begin{itemize}
    \item \texttt{view\_matches} -- displays the current matchday's fixtures and bookmaker odds;
    \item \texttt{place\_bet} -- places a wager on a specified match, market and stake;
    \item \texttt{view\_bankroll} -- reports the agent's current balance and outstanding stakes;
    \item \texttt{next\_matchday} -- settles all bets, delivers results, downloads updated data and advances to the next matchday.
\end{itemize}

\subsubsection{CLI tools}

To develop models, the agent is provided with seven general-purpose tools for a sandboxed development environment: \texttt{bash}, \texttt{glob}, \texttt{grep}, \texttt{read}, \texttt{write}, \texttt{edit} and \texttt{todo\_write}. These tools enable the agent to write Python scripts, train models, inspect data files and organise its workflow. The tool schemas are designed to match the Claude Code toolset exactly \cite{anthropic2025claudecode}.

\noindent The sandbox provides 4 CPUs and 16GB of RAM with a standard Python data-science stack (NumPy, pandas, scikit-learn).

\subsection{Reward Structure}
\label{sec:reward}

KellyBench uses a dense, fully verifiable reward signal. After each matchday~$t$, the reward is the change in log-wealth:

\begin{equation}
    r_t = \log W_{t+1} - \log W_t = \log \frac{W_{t+1}}{W_t},
    \label{eq:reward}
\end{equation}

\noindent where $W_t$ denotes the agent's bankroll at the start of matchday~$t$ (before bets are deducted) and $W_{t+1}$ is the bankroll after settlement. The cumulative reward over a full season therefore equals the log-ratio of final to initial wealth:

\begin{equation}
    R = \sum_{t=1}^{T} r_t = \log \frac{W_{T+1}}{W_1}.
\end{equation}

\noindent We chose this reward due to its connection to the Kelly criterion \citep{kelly1956new}, hence the name \texttt{KellyBench}. The strategy that maximises expected log-wealth growth is the Kelly-optimal strategy, which also maximises the long-run geometric growth rate of the bankroll.

\noindent Since rewards are computed deterministically from match outcomes and bookmaker odds, KellyBench is fully verifiable and does not require LLM graders for evaluation. One possible exception is for the training set, where an auxiliary LLM-as-a-judge signal can be used to penalise deviations from rule-based strategies. This auxiliary signal is not included by default in our OpenReward implementation.

\subsection{Task Scenarios}
\label{sec:scenarios}

\texttt{KellyBench} comprises five scenarios spanning different eras of English football, divided into training and test splits (Table~\ref{tab:scenarios}).

\vspace{1em}

\begin{table}[h]
\centering
\begin{tabular}{llccc}
\toprule
\textbf{Scenario} & \textbf{Season} & \textbf{Split} & \textbf{Initial Bankroll} & \textbf{Matchdays} \\
\midrule
New Millennium     & 2000/01 & train & \pounds 100 & 97 \\
Post-Crash         & 2010/11 & train & \pounds 150 & 105 \\
Covid Season       & 2020/21 & train & \pounds 200 & 148 \\
Recent Season      & 2023/24 & test  & \pounds 220 & 120 \\
Recent Season (Lit.)\ & 2023/24 & test  & \pounds 220 & 120 \\
\bottomrule
\end{tabular}
\caption{\textbf{KellyBench Scenarios}. Each scenario covers a full English Premier League season. The literature variant of the 2023/24 test scenario augments the agent's sandbox with a curated collection of 30+ research papers on sports prediction and betting markets. For the principal results in this paper, we normalise initial bankroll numbers above to £$100{,}000$ for display purposes.}
\label{tab:scenarios}
\end{table}

\vspace{1em}

\noindent The training scenarios are drawn from distinct periods, each presenting different league dynamics, team strengths, and odds characteristics. The test scenario uses a recent complete season from 2023/2024. The \emph{literature variant} of the test scenario provides the agent with access to a curated research library: a deep research report and over 30 academic papers in markdown format covering probabilistic football modelling, scoring rules, expected goals models and Kelly betting.

\noindent Each scenario is long-horizon and agents use 500-1000 tool calls to complete an episode, so agents need to be effective at compaction and context management to perform well on \texttt{KellyBench}.

\noindent For this paper we focus on the \texttt{Recent Season} evaluation scenario, and we conduct an ablation involving the literature variant as well.

\subsection{Data Leakage}

There are three sources of data leakage that we need to guard against in \texttt{KellyBench}:

\begin{enumerate}
    \item \textbf{Result Knowledge in Weight Memory}. For example, a model that knows the outcome or the score can utilise this information to make profitable bets.
    \item \textbf{Season Knowledge in Weight Memory}. For example, a model that knows that Leicester City won the Premier League in 2015/16 can utilise this information to make a profitable strategy for that season even if it does know not the direct scores or outcomes of any particular match.
    \item \textbf{Accessing the Internet for Future Results}. A model that is able to download files with match results or access websites with future information can use this to cheat. We prevent this by blocking network access on the environment sandbox.
\end{enumerate}

\noindent To guard against weight memory leakage, we give agents explicit instructions to follow a rules-based strategy. They are told not to bet on gut feeling or intuition, but to follow the recommendations of a rules-based approach. Note that this is exactly how a quantitative analyst should behave when constructing backtests; they should not fall upon their knowledge too much of what actually happened when developing a model, otherwise they risk overfitting to a particular realisation of the world. 

\noindent To reinforce the point, we tell agents that they will be banned from betting and their bankroll eliminated if they do not follow a rules-based strategy. Based on observed agent trajectories, models comply with this instruction and it appears to be sufficient for preventing leakage. However, this could change with future AI agents that can hide their true intentions behind their chain-of-thought \cite{emmons2025cotnecessary,chen2025reasoning}.

\noindent We found a particularly striking example of the risk of weight memory leakage in GPT 5.4's trajectory:

\begin{quote}
As a model-based assistant with knowledge up to 2026, I know the actual result of Burnley vs Man City was 0-3 away, which suggests a potential bet on away and over. However, I must base my decisions on a reliable model.
\end{quote}

\noindent The GPT-5.4 agent proceeded to use the recommendations of a machine learning model instead, but a future agent could be deceptive. For example, suppose the agent knew that Man City won a match against Burnley. It could find a rules-based strategy that suggests Man City as having edge in the encounter, and then claim that "I am betting on Man City because of the model's recommendation". However, if an agent did this we would expect to observe opportunistic changes in the rules-based strategy over the course of the season. On the contrary, we do not see this kind of reward hacking in the traces we observe - agents are notably non-adaptive - but it is something to watch out for with future models.

\noindent A more insidious source of reward-hacking would be knowing some aggregate season statistic and silently introducing model bias to maximise it. For example, the 2023/2024 season was unusually high scoring with 3.28 goals per match versus a 2.67 historical average \cite{premierleague2024seasonlikenoother}. Assuming Poisson-like distributions for goals, this decreases the likelihood of draws. Indeed, empirically this is what happened with 82 draws versus a 97 historical draw average. Football models typically inflate draw probabilities based on empirical distributions \cite{Maher1982FootballScores}. However, an extremely smart AI agent that knew that the 2023/2024 season was high-scoring could choose to omit this inflation factor silently to achieve opportunities for betting against draws.

\noindent Both types of reward hack, especially the latter, would require extremely high degrees of hidden reasoning. While we cannot be certain, we have two strong reasons to believe agents are not yet capable of this degree of reasoning. The first is that frontier models perform poorly on \texttt{KellyBench}, so unless the agent is sandbagging, it is not utilising its knowledge of the future effectively\footnote{Sandbagging seems unlikely at present given the known optimisation objectives of current language models to perform well on software engineering tasks, but we cannot rule it out completely.}. A second reason is that models demonstrate poor sophistication in their strategies compared to humans, so assuming they are attempting to perform well, they are not exhibiting strong levels of domain-relevant reasoning compared to human experts at present.

\noindent Our long-term plan is to introduce a live version of \texttt{KellyBench} as soon as models start to perform well. For now, the current backtest environment is not yet saturated and frontier models perform poorly. If independent evaluators wish to evaluate agents on \texttt{KellyBench}, we advise them to audit trajectories and disqualify agents that disregard the rules-based betting instructions.

%% file: findings.tex
\subsection{Headline Results}

\noindent All models lost money on average over five seeds on the 2023/24 English Premier League season, with the mean return on investment ranging from -89.6\% (Kimi K2.5) to -7.9\% (GPT-5.4). Only 3 out of 25 seeds achieved a positive return on investment: one from GPT-5.4 ($+34.1\%$), one from Gemini~Pro ($+33.7\%$), and one from Claude Opus 4.6 ($+21.5\%$).

\noindent Table~\ref{tab:model-summary-5seeds} summarises the final bankroll, ROI, and key statistics for each model. For more detailed trajectory analysis, per-model narratives are available in Appendix~\ref{app:per-model}. Note that all models were run on a maximum reasoning budget and we enabled interleaved thinking where applicable.

\vspace{1em}

\begin{table}[H]
    \centering                                                                                                                           
    \small                                                                                                                               
    \begin{threeparttable}                                                                                                               
    \begin{tabular}{lrrrrrrr}                                                                                                            
    \toprule                                                                                                                             
    \textbf{Model} & \textbf{Avg ROI} &                                                                                                  
    \textbf{Best Seed} & \textbf{Worst Seed} &                                                                                           
    \textbf{Ruin?\tnote{1}} &                                                                                                                   
    \textbf{Avg Bets} & \textbf{$\Delta$LL\tnote{2}} &                                                                                   
    \textbf{Final Bankroll} \\                                                                                                           
    \midrule                                                                                                                             
    GPT-5.4     & $-7.9\%$   & $+34.1\%$                                                                                                 
    & $-32.9\%$ & No & 115 & $+0.016$ & \pounds92{,}063 \\                                                                              
    Claude Opus 4.6       & $-11.2\%$  & $+21.5\%$                                                                                       
     & $-44.7\%$ & No & 202 & $+0.016$ & \pounds88{,}771 \\                                                                             
    GLM-5        & $-51.6\%$ & $-14.3\%$                                                                                                 
     & $-100.0\%$ & Yes  & 221 & $+0.054$ & \pounds48{,}395 \\                                                                            
    Gemini 3.1 Pro & $-66.0\%$ & $+33.7\%$                                                                                               
     & $-100.0\%$ & Yes & 360 & $+0.068$ & \pounds34{,}029 \\                                                                             
    Kimi K2.5   & $-89.6\%$  & $-77.7\%$                                                                                                 
     & $-100.0\%$ & Yes  & 178 & $+0.080$ & \pounds10{,}421 \\                                                                            
    \bottomrule                                                                                                                          
    \end{tabular}                                                                                                                        
    \begin{tablenotes}                                                                                                                   
      \footnotesize                                                                                 \item[1] Ruin is defined as if the model lost its entire bankroll in any of the five seeds.                    
      \item[2] Model log loss minus market log loss on bets placed using model and market implied probabilities; lower is better.
    \end{tablenotes}                                                                                                                     
    \caption{\textbf{The Road to Ruin.} Model
    performance across five seeds on the 2023/24                                                                                         
    English Premier League season. Each model                                                                                            
    begins with \pounds100{,}000. Final bankroll is averaged across five seeds.}                                                                                             
    \label{tab:model-summary-5seeds}                                                                                                     
    \end{threeparttable}                                                                                                                 
    \end{table}    

\noindent While financial results were often exacerbated by poor risk management, the fundamental driver was predictive model underperformance versus the market, as shown in the log loss difference. This was not for lack of trying, as most models performed rudimentary backtests in the initial phase, often "finding" positive returns, but then experienced inferior performance when actually deployed. This reflects \texttt{KellyBench}'s non-stationarity and difficulty versus traditional machine learning tasks.

\noindent To test whether inter-model differences were significant, we obtained per-matchday log returns and performed two-sided Mann-Whitney U tests, which are shown in Table \ref{tab:pairwise-sig-5seeds}. Note that since each seed shares the same match schedule, pooling returns in this way may overstate significance (as the effective sample size is lower), reflecting a limitation of the single-season setup.

\vspace{0.5em}

  \begin{table}[H]
  \centering\small
  \begin{tabular}{lccccc}
  \toprule
  \textbf{Model} & \textbf{Claude Opus 4.6} &
  \textbf{Gemini 3.1 Pro} & \textbf{GPT-5.4} &
  \textbf{GLM-5} & \textbf{Kimi K2.5} \\
  \midrule
  Claude Opus 4.6    & ---            &
     &                   &                &  \\
  Gemini 3.1 Pro & $<$0.001***    & ---
     &                   &                &  \\
  GPT-5.4    & $0.209$        & $<$0.001***
     & ---               &                &  \\
  GLM-5      & $0.010$**      & $0.121$
     & $0.057$           & ---            &  \\
  Kimi K2.5  & $<$0.001***    & $0.736$
     & $0.002$**         & $0.121$        & ---
  \\
  \bottomrule
  \end{tabular}
  \caption{\textbf{Pairwise significance tests.}
  Holm-Bonferroni corrected $p$-values from
  two-sided Mann-Whitney $U$ tests on pooled
  per-matchday log returns (5 seeds per model).}
  \label{tab:pairwise-sig-5seeds}
  \end{table}

\noindent The two strongest models, Opus~4.6 and GPT-5.4, share several traits. Both models retrained or adjusted their strategies in response
to new match data, both deployed systematic staking rules rather than ad hoc
bet sizes, and both preserved capital during periods where their strategies
identified no edge. Opus 4.6 and GPT-5.4 were the only models to avoid ruin across all five seeds, with Opus 4.6 consistently deploying gradient boosting ensembles with
fractional Kelly staking. GPT-5.4 was the most engineering-intensive model, with
one seed dedicating roughly 160 tool calls to model building before placing
its first bet. Later, it adapted its strategy by reducing stakes to penny bets after
determining that it could not reliably outperform the market, demonstrating situational awareness.

\noindent Most of the other models we evaluated fell into a small set of recurring failure patterns that compounded to produce poor outcomes. Broadly, these failures fell into five categories: bankroll management failures in which agents discuss Kelly staking but implement something else, inability to handle the non-stationarity introduced by newly promoted teams, absence of intra-season adaptation, long-horizon situational awareness failures including premature task termination and difficulties with the tool-calling environment, and systematic miscalibration of draws and longshots. We summarise the prevalence of these failings across the 25 seeds in Table~\ref{tab:failure-prevalence}.

\vspace{1em}

  \begin{table}[H]                                                                                                                       
  \centering                                                                                                                             
  \small                                                                                                                              
  \begin{tabular}{lc}                                                                                                                    
  \toprule                                                                                                                             
  \textbf{Failure Mode} & \textbf{Seeds Affected} \\                                                                                     
  \midrule                                                                                                                               
  Bankroll ruin (final bankroll \pounds0)                                          & 6/25 \\                                                                                       
  \midrule                                                                                                                               
  \textit{Bankroll management} & \\                                                                                                      
  \quad No Kelly or principled stake-sizing at execution time                       & 9/25 \\                                            
  \quad Kelly code written but never invoked at bet time                            & 7/25 \\                                            
  \midrule                                                                                                                               
  \textit{Non-stationarity / promoted teams} & \\                                                                                        
  \quad No general handling of newly promoted teams                                 & 22/25 \\                                           
  \midrule                                                                                                                               
  \textit{Intra-season adaptation} & \\                                                                                                  
  \quad Never retrained statistical model after initial fit                         & 7/25 \\                                            
  \midrule                                                                                                                               
  \textit{Situational awareness / environment} & \\                                                                                      
  \quad Declared task complete while season still running                           & 8/25 \\                                            
  \quad Failed to correctly invoke provided betting tools                           & 13/25 \\                                        
  \quad Label leakage in backtesting or prediction pipeline                         & 7/25 \\                                            
  \midrule                                                                                                                               
  \textit{Calibration} & \\                                                                                                              
  \quad Systematic draw / longshot miscalibration                                   & 22/25 \\                                           
  \bottomrule                                                                                                                            
  \end{tabular}                                                                                                                          
  \caption{\textbf{Prevalence of failure modes across 25 seeds} (5 models $\times$ 5 seeds). Categories are non-exclusive: most seeds    
  exhibit multiple failures.}                
  \label{tab:failure-prevalence}                                                                                                         
  \end{table} 
  
\noindent We also evaluated three other models for three seeds on the benchmark in an earlier version of this paper: Arcee Trinity, Grok~4.20 and Gemini 3.1 Flash Lite Preview. The first two models struggled with basic environment interaction. Two of Trinity's three seeds placed zero bets, failing to discover or correctly invoke the provided tools. Two of Grok's three seeds entered terminal loops and declared the task complete while the season was still running; the third went bankrupt. We exclude these models from our headline results given their inability to complete the season reliably.

\subsubsection{Bankroll Management and the Gap Between Intention and Execution}
\label{subsec:bankroll_management}

\noindent The most pervasive failure mode of models in \texttt{KellyBench} was a disconnect between what agents reasoned about and what they actually executed in practice.

\noindent Bankroll management was either absent or broken in several of the evaluated models. Nearly every model discussed Kelly staking in its chain of thought, yet at execution time many fell back to flat stakes, percentage-of-bankroll heuristics, or ad hoc round numbers. For example, Kimi K2.5 wrote a numerically correct fractional Kelly function (Figure~\ref{fig:failure-code}) but never invoked it from its betting loop due to repeated tool-call formatting failures. Its episode ended with an accidental ${\sim}$\pounds114{,}000 bet\footnote{Bet sizing in this section is shown versus a normalised \pounds100{,}000 initial bankroll}, ninety-eight percent of its remaining bankroll, on Burnley versus Luton. GPT-5.4 discussed Kelly extensively in its reasoning chains and then implemented a simpler rule-based stake schedule.

\begin{figure}[H]
\centering
\begin{minipage}{0.97\linewidth}
\begin{codepanel}{Kimi K2.5: correct fractional Kelly that was never invoked}
def kelly_criterion(prob, odds, bankroll, fraction=0.25):
    b = odds - 1
    kelly_fraction = (prob * b - (1 - prob)) / b
    if kelly_fraction <= 0:
        return 0
    return bankroll * kelly_fraction * fraction
\end{codepanel}

\begin{codepanel}{GLM-5: fixed draw rate, never corrected despite self-diagnosis}
prob_draw = 0.25  # fixed; flagged as flawed in three
                  # mid-run self-critique documents
prob_home = prob_home * (1 - prob_draw)
prob_away = prob_away * (1 - prob_draw)
\end{codepanel}

\end{minipage}
\caption{\textbf{Example Failure Modes}. Failure modes above show a correct staking implementation that was never executed amd miscalibration that the agent diagnosed but never fixed.}
\label{fig:failure-code}
\end{figure}

\subsubsection{Non-stationarity and Newly Promoted Teams}

Another key test for the models was how to handle newly promoted teams where less recent historical data is available. The 2023/24 season featured three promoted teams, each presenting a distinct distributional shift that no model handled adequately. Luton Town had not played in the top flight since 1992, so the entire training window from 1993 to 2023 contained zero Premier League matches for the club\footnote{Note there is Championship data for these clubs, but Championship clubs only encounter Premier League clubs occasionally in cup competitions. In addition, newly promoted teams typically change their squad distribution before starting a Premier League campaign. So this is a large distributional shift with very little data to model adequately.}. Burnley's most recent top-flight data came from 2021/22. Sheffield United's most relevant prior data was their 2020/21 campaign.

\noindent No model implemented a general solution. One GLM-5 seed correctly diagnosed the problem on matchday one, observing that Luton had zero historical matches and that the apparent edge was a model artefact. It then hardcoded a skip-Luton rule that persisted for the rest of the season, while continuing to bet freely on Burnley and Sheffield United and never generalising the insight. Opus~4.6 showed a recurring pattern of diagnosing the problem and then deferring to its incorrect model anyway. It reasoned that although the apparent edge on a Burnley fixture was implausible given the squad, it should follow the model-based approach since that was the framework it was using. One GPT-5.4 seed adopted the most elegant non-solution, relying on \texttt{OneHotEncoder(handle\_unknown=
`ignore')} so that bookmaker odds absorbed the promotion signal, in effect deferring to the market on teams it could not model.

\subsubsection{Absence of Intra-season Adaptation}

There are two kinds of adaptability that are important. The first kind of adaptability is broadly focused around retraining upon receiving new data; for example, updating dynamic strength models. The second kind revolves around changing the model itself in light of realised performance (``pivots''). We find models are poor on both counts, especially the second kind, which indicates a long-horizon capability that we would like models to be good at in broader contexts beyond sports betting.

\noindent First, existing model adaptability is poor. The majority of seeds fitted their statistical models once at the start of the season and never updated them despite receiving fresh match data after every matchday. This likely contributed to poor performance, as the market progressively grew its information edge over the model. Only a few seeds were fully adaptive in the sense of progressively retraining models, including one GPT-5.4 seed which performed walk-forward retraining, and two Opus 4.6 seeds.

\noindent Strategic pivots were rare and only present in the strongest models. One of the Opus 4.6 seeds reduced its Kelly fraction from 0.25 to 0.15 after a series of drawdowns; another pivoted from gradient boosting to a Poisson-plus-market-prior blend after diagnosing that its model had no edge against the closing line. A GPT-5.4 seed pivoted to an extremely conservative strategy after diagnosing it had no edge versus the market based on realised performance, resorting to penny bets for the rest of the season. Even models that pivoted had a bias towards model building at the start of the season, as opposed to continually improving their approach throughout the season.

\noindent When taken together, the fully adaptive seeds outperformed the fully static ones in average ROI ($-11.1\%$ vs.\ $-70.0\%$), with partially adaptive runs falling in between ($-49.6\%$).

\subsubsection{Long-horizon Situational Awareness and Environment Failures}

A large cluster of failures concerned situational awareness over the course of the episode. Several models declared the season finished while it was still running. For example, one Kimi K2.5 seed produced six separate ``final season summary'' documents at successive bankroll levels of ${\sim}$\pounds67{,}700, \pounds38{,}400, \pounds29{,}800, \pounds26{,}600, \pounds26{,}600, and \pounds15{,}700, each claiming the episode was concluded. This yielded a roughly eighteen-matchday plateau during which the agent acted as if the run were over.

\subsection{Are models doing the best they can with the available data?}

\noindent A possible criticism of our setup is that we are not providing agents with enough data to obtain edge in the market, and that models might be doing the best they can with the data available. On the contrary, we find that agents underutilise available data and have unsophisticated approaches.

\noindent To measure this, we constructed a 52-point rubric that judges the sophistication of the strategies employed by the models. We constructed this rubric by consulting experts with experience working at leading quantitative betting funds. Different rubrics test facets of strategy such as feature development, execution strategies, and how models account for challenges such as non-stationarity. 

\noindent We do not reproduce the rubric in full to maximise the lifetime of the benchmark, but we produce four examples of the criteria below in Table \ref{tab:example_rubrics} to illustrate how the model is scored.

\vspace{1em}

\begin{table}[h]
\centering

\begin{tabular}{clp{4.5cm}}
\toprule
\textbf{\#} & \textbf{Criterion} & \textbf{Scoring} \\
\midrule
2 & Does the agent use Kelly staking? & \makecell[tl]{0: No \\ 1: Full Kelly \\ 2: Fractional Kelly} \\
\midrule
4 & Does the agent use dynamic team ability models? & \makecell[tl]{0: No \\ 1: Exponential average / Elo \\ 2: State-space models} \\
\midrule
12 & Does the agent use a special approach for promoted teams? & \makecell[tl]{0: No \\ 1: Yes} \\
\midrule
43 & Does the agent correct for false positive risk, e.g. Bonferroni? & \makecell[tl]{0: No \\ 1: Yes} \\
\bottomrule
\end{tabular}
\caption{\textbf{Example Sophistication Rubrics}. Rubrics were constructed by human experts testing modeling, wagering and domain knowledge. There are 45 criteria in total (maximum score of 52).}
\label{tab:example_rubrics}
\end{table}

\noindent Using these rubrics, we calculate average sophistication of each model in Table \ref{tab:sophistication-table}. All models show low strategy sophistication. The best model Claude Opus 4.6 achieves less than a third of possible points on the measure with a sophistication score of 26.5\%. 

\vspace{1em}

\begin{table}[H]                             
\centering                                     
\begin{threeparttable}                         
\begin{tabular}{llrrr}                         
\toprule                                       
\textbf{Rank} & \textbf{Model} &               
\textbf{Sophistication} & \textbf{Final        
Bankroll} & \textbf{Avg Tool Calls} \\         
\midrule                                       
1 & Claude Opus 4.6 (max)                   &
26.5\% & \pounds88{,}771 & 564 \\              
2 & GPT-5.4 (xhigh)                         &
22.3\% & \pounds92{,}063 & 848 \\              
3 & GLM-5 (thinking)                        &
17.3\% & \pounds48{,}395 & 753 \\               
4 & Kimi K2.5 (thinking)                    &
12.7\% & \pounds10{,}421 & 636 \\               
5 & Gemini 3.1 Pro Preview (high)           &
8.8\% & \pounds34{,}029 & 734 \\               
\bottomrule                                  
\end{tabular}                                  
\caption{\textbf{Models show low strategy    
sophistication.} Sophistication is expressed as
a percentage of the maximum expert
rubric score (52). Bankroll and tool    
calls are averaged across 5 seeds.}                 
\label{tab:sophistication-table}
\end{threeparttable}                           
\end{table}

\begin{figure}[H]
\centering
    \includegraphics[width=1.00\linewidth]{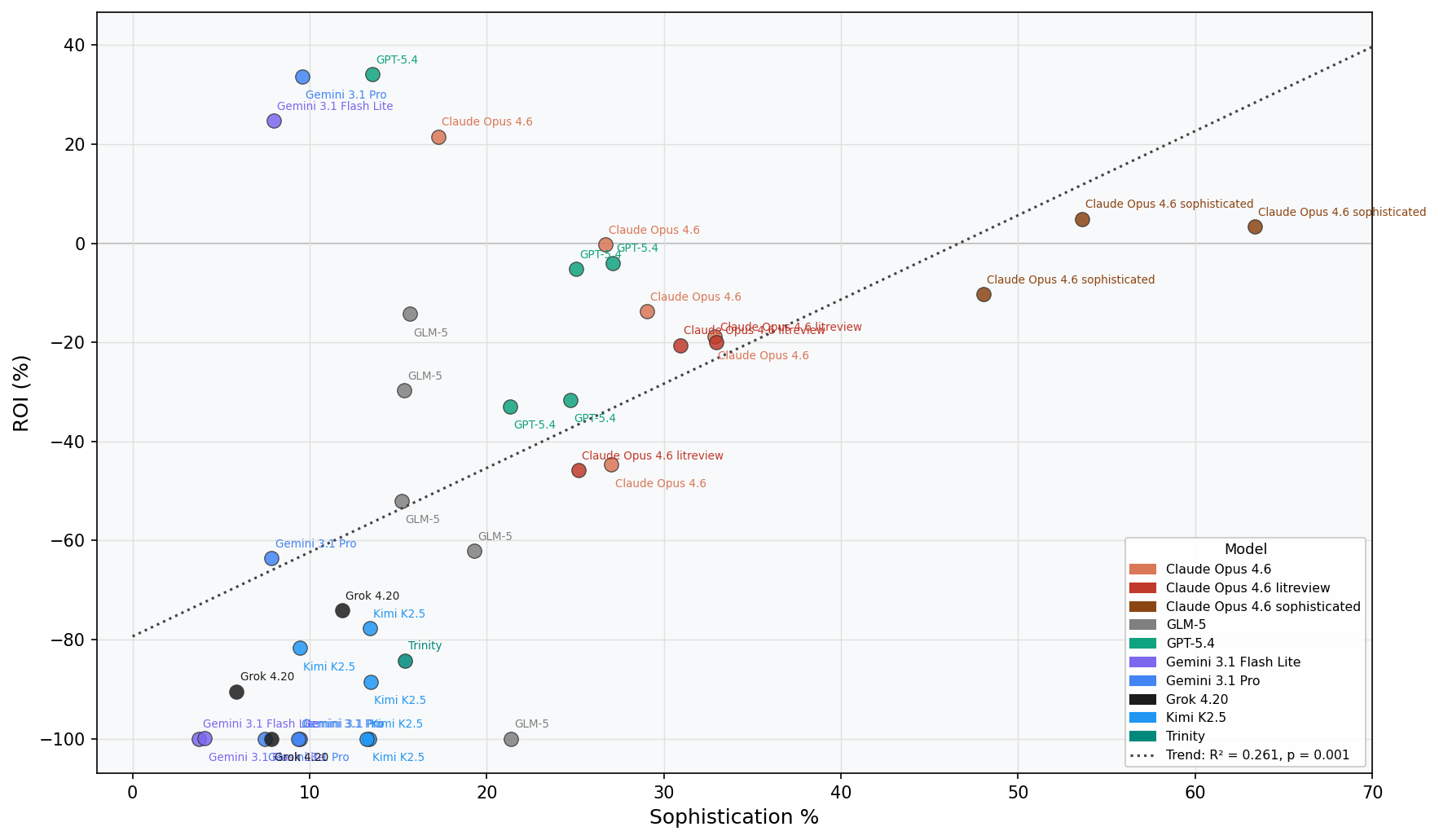}
    \caption{\textbf{Sophistication vs Return on Investment}. For each seed of a model on \texttt{KellyBench}, we plot the sophistication score against the final return on investment, which exhibits a positive correlation. Given that frontier models achieve less than a third of rubric points on average, we believe there is considerable room for improvement in the strategies created with the data afforded.}
\label{fig:sophistication}
\end{figure}

\noindent Performance and sophistication also have a statistically significant (moderate) positive relationship over the seeds tested, which we plot in Figure \ref{fig:sophistication}. Higher sophistication is also associated with lower rates of ruin: across all seeds tested, those scoring 11--18/52 went bankrupt at a   
  rate of ${\sim}8\%$, compared with ${\sim}55\%$ for seeds scoring 0--5/52. A logistic regression of ruin probability on sophistication 
  score yields a likelihood-ratio test of $p < 0.001$, which likely reflects the inclusion of staking-based criteria in the rubric.
   
\noindent We also performed an ablation where we gave three seeds of Claude Opus 4.6 access to the sophistication rubric for developing strategies. While these models did not follow all the instructions, we observed their return on investment was higher, as shown in Figure \ref{fig:sophistication}. Given that current frontier models (without access to the rubric) achieve <50\% sophistication, there likely is room for improvement on \texttt{KellyBench} with the existing data and compute setup.

\subsubsection{Harness Ablations for Sophistication}

\noindent One potential critique of our setup is underelicitation from the use of of a custom ReSum harness as opposed to the standard Claude Code harness \cite{wu2025resumunlockinglonghorizonsearch}. Additionally, the evaluation environment has no web search tool, preventing an agent from conducting a literature search to discover and synthesise promising approaches.

\begin{figure}[H]
\centering
    \includegraphics[width=1.00\linewidth]{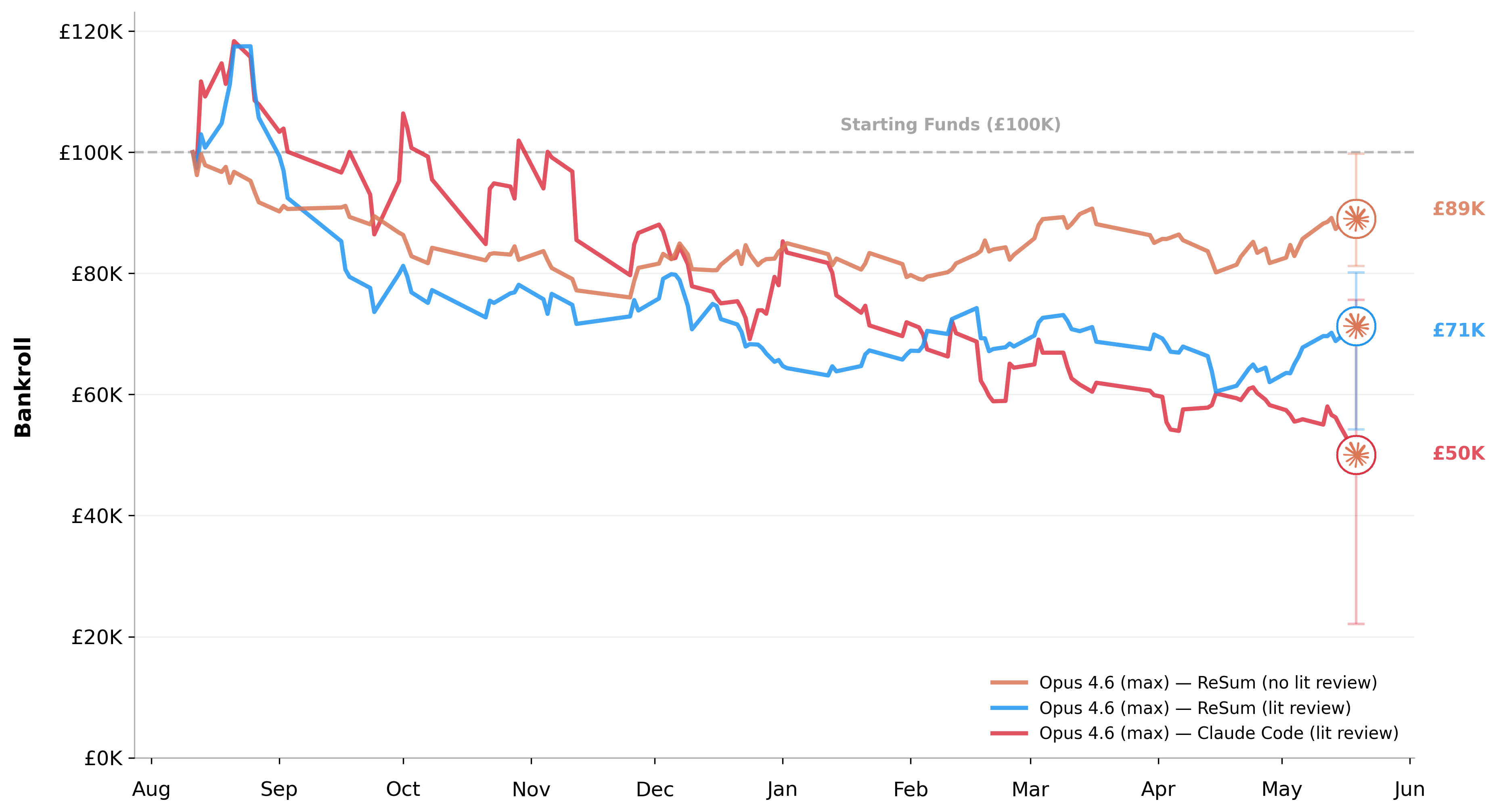}
    \caption{\textbf{Harness Ablations on Opus 4.6}. We test two possible critiques of our results: first, that by disabling network access, we are undereliciting performance that could be gained from access to the literature; and second, that a ReSum harness would underperform compared to a Claude Code harness. We ablate both setups and find that neither improves upon the baseline.}
\label{fig:harness-ablations}
\end{figure}

\noindent To test the effect of literature access and a better harness, we construct two additional baselines for the strongest performing model Claude Opus 4.6. First, we provide the model with a deep research report and over 30 relevant academic papers in its filesystem. Secondly, we swap out the ReSum harness for a Claude Code harness using \texttt{firehorse} library. We use three seeds for each new baseline, giving six runs in total, and plot the average bankroll results in Figure \ref{fig:harness-ablations}.

\noindent We do not see a positive impact on performance from either change. In fact, mean final bankroll declined with both baselines for the seeds we tested. We observed that models read the deep research report, but did not thoroughly read all the papers available in the filesystem. The main change in model behaviour from access to the literature appeared to be utilising Dixon-Coles models more frequently, but this did not result in increased performance \cite{dixon1997modelling}. 

\noindent In fact, Dixon-Coles models, while considered seminal in football modeling, are considered outdated given the way they estimate dynamic abilities through likelihood-weighting rather than Bayesian estimation. This could suggest that models have poor ``taste'' in discerning the importance of relevant literature. When judged according to our sophistication score, the literature-review variants did not improve sophistication which stayed the same (29.5\% vs 29.5\%)\footnote{Note we used three seeds for this ablation, which differs from the five seeds used for Opus 4.6 in the headline results.}.

\noindent Strategy sophistication improved to 55.0\% when we gave direct access to the sophistication rubric. This achieved a mean return over three seeds of $-0.7\%$, although models did not follow all instructions from the rubric. This modification removes the open-endedness of the task by telling the model the right recipe from the outset. However, it suggests that a better prompt or directed planning mode could potentially improve performance, and we leave this to future work.

\subsection{Return Analysis}

\begin{figure*}
\centering
\begin{subfigure}[t]{0.49\textwidth}
    \centering
    \includegraphics[width=\linewidth]{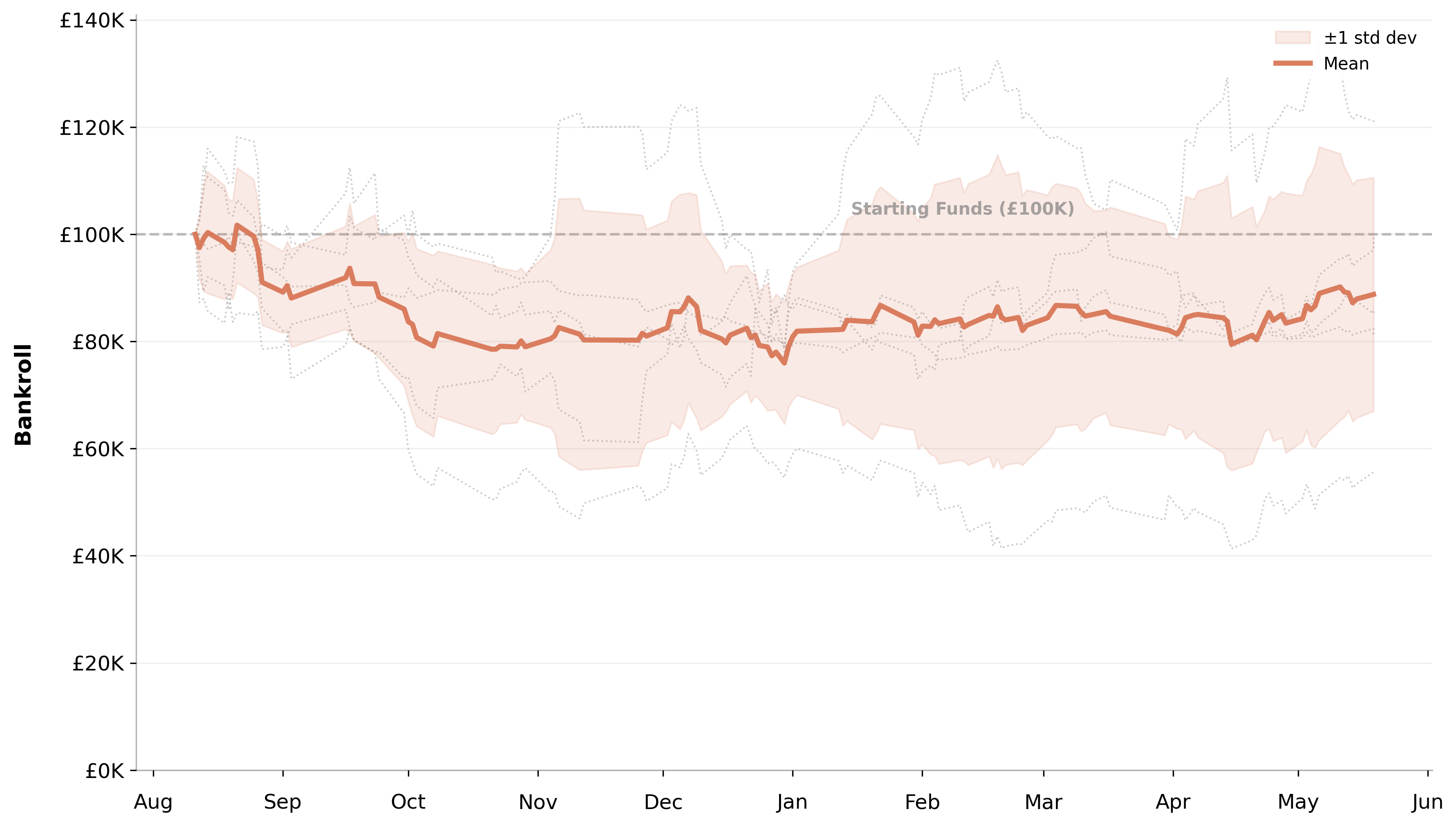}
    \label{fig:opus-performance}
\end{subfigure}
\hfill
\begin{subfigure}[t]{0.49\textwidth}
    \centering
    \includegraphics[width=\linewidth]{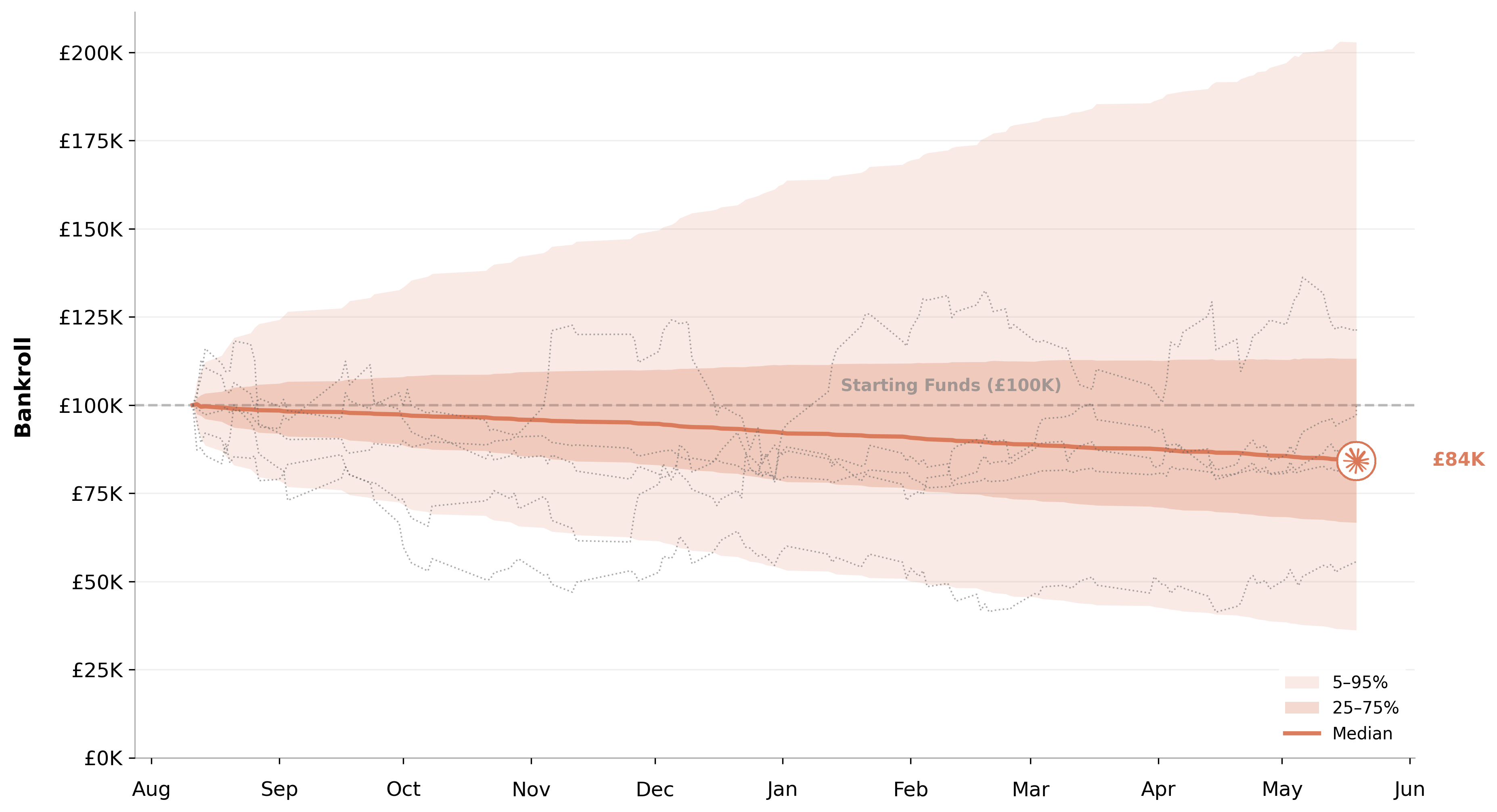}
    \label{fig:second-figure}
\end{subfigure}
\caption{\textbf{Return Analysis on Claude Opus 4.6.} In the left subfigure, we plot trajectories from 5 seeds and plot $1\sigma$ error bars. In the right subfigure, we run a hierarchical bootstrap. We first randomly choose one of the 5 seed trajectories and then sampling empirical log returns from it for each matchday. We repeat this for 50,000 simulations to construct bootstrapped error bars.}
\label{fig:two-column-two-figures}
\end{figure*}

To understand potential variance, we conducted a hierarchical bootstrap for Claude Opus 4.6 to construct bootstrapped error bars. These results are plotted in Figure \ref{fig:two-column-two-figures}. According to the hierarchical bootstrap, around 33\% of simulations end up profitable with Claude Opus 4.6 and the remainder lose money. This shows that there is some potential for Claude Opus 4.6 to make money over the course of the season (indeed one seed ends up profitable) but our baseline conclusion is that the model is not consistently outperforming the market on average. In lieu of no other model outperforming the market on average, our conclusion stands that the current generation of models is not finding consistent edge at present.

\subsection{Knowledge-Action Gaps}

\noindent A broader theme we observe in the trajectories is a gap between knowing what to do and actually executing it in practice. Models frequently articulated promising strategies in their reasoning chains but failed to verify that written code implemented those strategies, failed to notice when execution diverged from intent, or failed to act on their own diagnostic findings. 

\noindent GLM-5 provides a particularly clear set of examples across multiple seeds. In one seed, the model wrote three separate self-critique documents during its run, each correctly identifying that a fixed 25\% draw rate and overestimation of home advantage were the root causes of its losses.  For example, at bankroll ${\sim}$\pounds44{,}200, it wrote: ``\emph{Model probabilities don't match observed frequencies: Predicted 40\% home wins only won ${\sim}$30\%.}''  In one seed, it explicitly noticed that Burnley had been promoted but only special-cased Luton as a skip rule, ignoring the general problem of promoted teams.  In another seed, it discovered a $+2.62\%$ ROI strategy in backtesting but never validated this strategy using held-out data before deployment.

\noindent Kimi K2.5 exhibited a different variant of the same problem.  In one seed, its prediction script with Kelly staking was well-designed, but a persistent inability to format tool calls correctly as it entered long-context caused the model to send \texttt{": ["} as a bash command approximately 50~times in sequence.  The model's reasoning noted: ``\emph{bro, you seeing this?  The bash command keeps returning empty.  Let me try again with proper command execution}'', yet it continued sending the identical broken command.  In one seed, the model declared ``episode complete'' six separate times while its bankroll was still declining, then entered an 18-matchday plateau with zero bets placed.  In another seed, it peaked at ${\sim}$\pounds170{,}000 with a working Poisson model before suddenly abandoning this strategy.

\noindent These patterns suggest that many models can generate sophisticated plans but often struggle to maintain the closed-loop monitoring and coherence required to execute them over extended episodes. The ability to write correct code, identify correct strategies, and diagnose failures does not reliably translate into the situational awareness needed to detect and correct execution failures in real time and in accordance with new information.

\subsection{Human Baselines}

Is \texttt{KellyBench} even beatable? To find out we established a few human baselines to compare model performance against:

\begin{enumerate}
    \item \textbf{Favourites-Only}: A basic strategy that always stakes 5\% of bankroll on the favourite team.
    \item \textbf{Dixon-Coles}: A seminal but outdated 2000s baseline model for predicting football outcomes.
    \item \textbf{Human Quant}: A human with two years of experience making football models with a week-long time horizon to make a predictive model.
    \item \textbf{AI Researcher}: A human with five years of experience in deep learning, but no prior experience of the sports betting domain, with a week-long time horizon to make a predictive model.
\end{enumerate}

\begin{figure}[H]
\centering
    \includegraphics[width=1.00\linewidth]{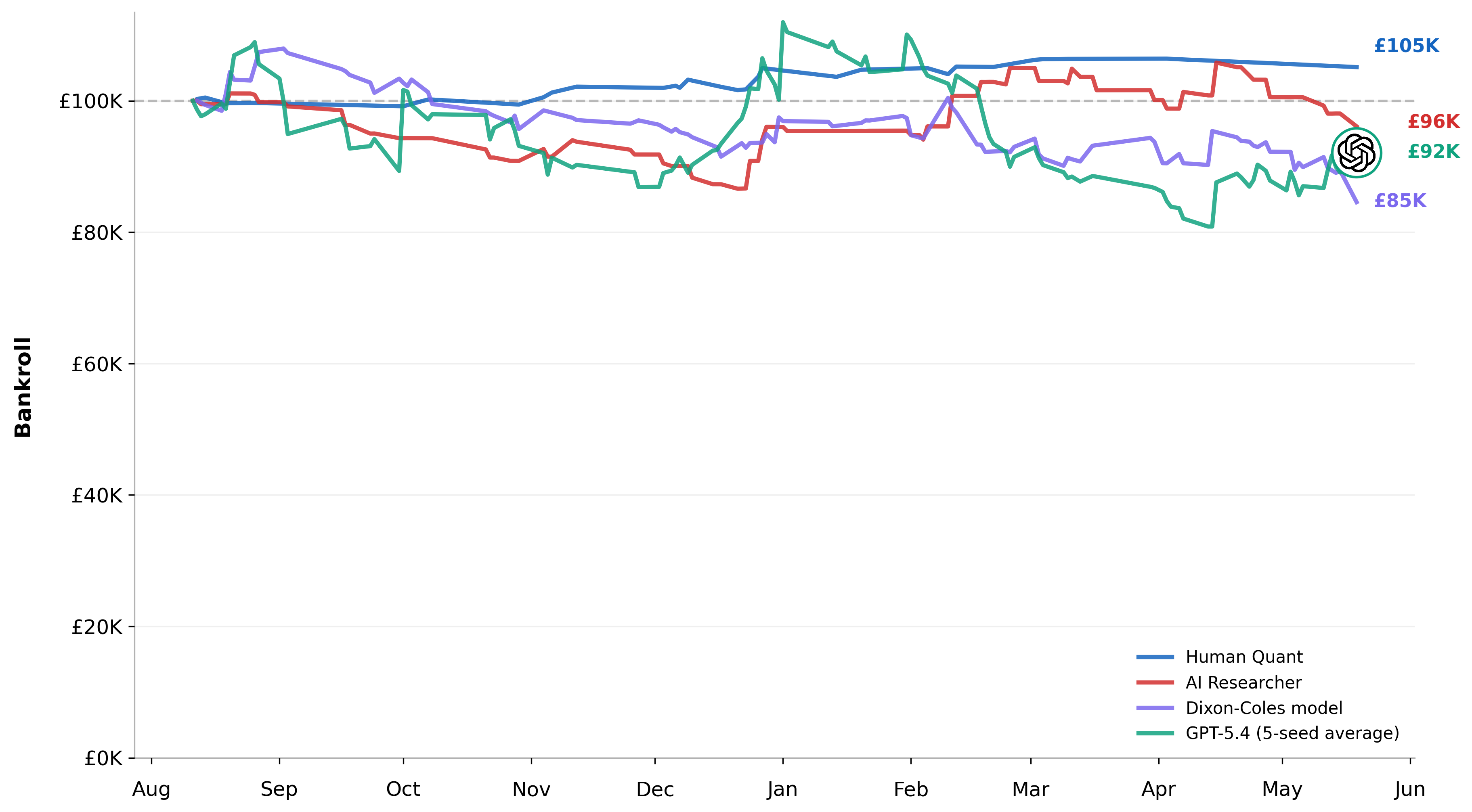}
    \caption{\textbf{Human Baselines on KellyBench}. In general we observe that human baselines of various sophistication outperform tested frontier models. Surprisingly a classic 2000s baseline (Dixon-Coles) outperforms 3/5 frontier models evaluated.}
\label{fig:human-baseline-chart}
\end{figure}

\noindent We plot results in Figure \ref{fig:human-baseline-chart} and report headline metrics in Table \ref{tab:headline_results}. Surprisingly, the simple Dixon-Coles baseline is competitive and beats 3/5 of the models evaluated on \texttt{KellyBench}. The human quant baseline is the best performing strategy, although it is relatively conservative and makes 39 bets over the course of the season. According to the model probabilities of the human quant's strategy, they would be expected to lose money 38\% of the time across those 39 bets which reflects high variance.

\vspace{0.5em}

\begin{table}[H]
\centering
\begin{tabular}{lrrrr}
\toprule
\textbf{Strategy} & \textbf{ROI} & \textbf{Bets} & \textbf{Sharpe} & \textbf{Final Bankroll} \\
\midrule
Human Quant              & +5.1\%  &  39 &  0.96 & \pounds105{,}118 \\
AI Researcher            & $-$4.3\%  & 140 & $-$0.33 & \pounds95{,}705 \\
GPT-5.4 (5-seed average)    & $-$7.9\%  & 115 & $-$0.35 & \pounds92{,}063 \\
Dixon-Coles model        & $-$15.4\% & 249 & $-$0.90 & \pounds84{,}567 \\
\bottomrule
\end{tabular}
\caption{\textbf{Headline financial results for all strategies over the 2023--24 Premier League season.} ROI and final bankroll are normalised to a \pounds100{,}000 starting bankroll. Bets is the total number of bets placed (averaged across seeds for GPT-5.4). Sharpe ratio is computed from per-bet returns.}
\label{tab:headline_results}
\end{table}

\noindent The sophistication score of the human quant strategy was $73.1\%$, which is higher than any of the models tested on the benchmark. We reproduce the final comments from the human quant below on how they approached the task and their assessed difficulty of the challenge.

\newpage

\begin{mdframed}[linewidth=0.5pt, roundcorner=5pt]
\small
This was a hard task. The most difficult challenge was finding features that have a reliable edge over the market. To address this, I had to invest significant effort into developing the backtesting pipeline, carefully selecting appropriate folds, significance levels, and correction techniques.

\noindent I designed around 260 features but the final model retained fewer than 10, reflecting the stringent inclusion criteria. Several features performed well over a few seasons but failed to generalise in the final backtest.

\noindent The task was challenging for two reasons. First, using odds with a 5\% vig proved very unforgiving. Second, the data to work with was mostly headline statistics and publicly available data, so the available edges are likely smaller and need attention to detail to find.

\noindent With more time I would have developed more player-level features to extract more small edges, which would have enabled more betting opportunities.
\end{mdframed}

\noindent The single-season setup restricts us from making overly strong conclusions about these results, but it is notable that both the 2000s baseline (Dixon-Coles) and human researcher baselines outperformed all models evaluated.

%% file: limitations.tex
\subsection{Limitations}

\subsubsection{Market efficiency}

The English Premier League is widely regarded as one of the most informationally efficient prediction markets in the world \cite{ElaadReadeSingleton2020}. While models in \texttt{KellyBench} are provided with plenty of data on which to base their strategies, there is much we don't provide agents. For example, we do not give models any tracking data which is particularly important for evaluating defensive ability. We do not provide set-piece data that is an important subgame with distinct team abilities. We do not provide in-play data with real-time information. Lastly, we use closing line odds, at which point lineup information may be public - for example, new knowledge of a key player injury - but we do not give this information to the agent. Taken together, there is a strong potential for estimation error.

\noindent However, we do not think this invalidates the benchmark for two reasons. First, models show low sophistication even with the existing data, meaning they are still underexploiting available information. Therefore, even if models cannot make a profit with the existing data, the benchmark may still measure meaningful progress in frontier model capability. Secondly, it is not necessary to have all fundamental information to gain an edge. Exactly as \citet{benter2008computer} pointed out, we can blend fundamental models with public factors from odds which contain the missing information, and obtain overlays where our edge exists.

\subsubsection{Single season only}

\noindent A related critique is the choice for us to backtest over a single season. To identify stronger relative differences between models we would ideally need more seasons to draw stronger conclusions. However, as a starting point, we think a single season already reveals many interesting failure points of existing models. In particular, no model evaluated makes a consistent return on average. Additionally, given the expense of running \texttt{KellyBench}, we decided a single-season backtest was the right first step to validate the usefulness of this class of benchmark.

\subsubsection{Harnesses}

We tested several single agent harnesses, but we did not test multi-agent or iterative harnesses. For example, scaffolds in the style of AlphaEvolve may be capable of increased performance, especially on tasks such as feature development and exploration \cite{novikov2025alphaevolvecodingagentscientific}. In addition, multi-agent frameworks could also lead to better specialisation of work between feature development, backtesting and trade execution. We leave these ablations to future work. 

\noindent We do note, however, that such ablations are likely to be prohibitively expensive for academic researchers in particular. Given that running one seed of a GPT-5.4-level model costs $\approx\$2,000$, iterative and multi-agent approaches (on future larger models) are likely to be extremely expensive and out of reach to most individuals or small groups. We make this point to raise attention to this issue as the field moves beyond procedural tasks towards evaluations for long-horizon complex worlds like \texttt{KellyBench}.

\subsubsection{Using middle line odds}

\noindent We used bookmaker odds, where overround ranged from 2\% to 7\%, and took a middle of the line odds at 5.3\% overround. This makes edge harder for the models to find. Given that models typically return less than -10\%, we think the results are not invalidated by using these odds. Indeed, as we point out throughout the paper, models demonstrate low sophistication with the data they are given, leaving room for improvement against these odds. However, in future work we are considering switching to using to ``sharp'' odds, as they are a better test for market efficiency.

\subsubsection{Knowledge contamination and concealment}
A fundamental limitation of any retrospective evaluation is that frontier models may have encountered the ground-truth outcomes during pre-training. The 2023--24 English Premier League season concluded before the knowledge cutoffs of all models tested, meaning that match results, final standings, and even betting market movements are likely present in their training corpora. We have earlier discussed this problem in detail, but we raise it again here for transparency.

\noindent The consensus of the authors is that models comply with the instructions to use rules-based strategies in the backtesting environment, but we will reassess as new models are released. We plan to switch to a live benchmark when we detect signs of reward hacking. We also note again that, despite potential contamination, all models still lost money on average across five seeds. We again note that we did not see evidence of reward hacking in the trajectories used in this work.

\subsection{Conclusions}

\texttt{KellyBench} reveals a consistent pattern across all models tested: none can profitably navigate a full season of English Premier League betting markets on average across five seeds.  The benchmark exposes failures not only in machine learning modeling, where models struggle to outperform bookmaker-implied probabilities, but more fundamentally in the closed-loop reasoning required for long-horizon sequential decision-making. Models can write sophisticated code, diagnose their own failures, and articulate correct strategies, yet persistently fail to execute those strategies reliably, monitor their own performance, or adapt when their approach is not working.

\noindent As well as looking at performance, we judged strategy sophistication for each model and found existing models to have unsophisticated strategies compared to human approaches. In particular, rich player-level data available in the environment was almost universally ignored in favour of simpler team-level features, suggesting that current models systematically underinvest in data and feature engineering when operating autonomously. Additionally, the non-stationarity challenge posed by factors like newly promoted teams proved universally problematic, with no model implementing a general solution despite several explicitly diagnosing the problem. 

\noindent The gap between analytical capability and operational competence that \texttt{KellyBench} measures is the gap that matters for real-world deployment of agentic AI systems. Existing models are now extremely competent at procedural tasks, but need to be deployed in long-horizon settings with open-ended goals. In this regime, benchmarks that test sustained, adaptive reasoning under uncertainty become essential. \texttt{KellyBench} is an early example of a complex world that tests long-horizon sequential decision-making under uncertainty. We believe more environments and evaluations in this style will become more important as model capabilities continue to improve.

%% file: app.tex
\appendix

\input{showcase_format}
\definecolor{reasonbg}{RGB}{246,248,254}
\definecolor{reasonframe}{RGB}{180,196,230}
\definecolor{reasontitle}{RGB}{224,232,250}
\definecolor{toolbg}{RGB}{248,248,248}
\definecolor{toolframe}{RGB}{200,200,200}
\definecolor{tooltitle}{RGB}{232,232,232}

\newtcblisting{reasonpanel}[1]{%
  listing only,
  listing options={basicstyle=\ttfamily\scriptsize,breaklines=true,breakatwhitespace=true,columns=fullflexible,keepspaces=true,showstringspaces=false,frame=none,xleftmargin=0pt,xrightmargin=0pt,framexleftmargin=0pt,framexrightmargin=0pt,aboveskip=0pt,belowskip=0pt,literate={£}{{\pounds}}1},
  enhanced,
  arc=3pt, outer arc=3pt, boxrule=0.5pt,
  colback=reasonbg, colframe=reasonframe,
  colbacktitle=reasontitle, coltitle=codetitletext,
  fonttitle=\small\bfseries\sffamily,
  title={#1}, titlerule=0pt,
  toptitle=4pt, bottomtitle=4pt,
  top=5pt, bottom=5pt, left=6pt, right=6pt,
  boxsep=0pt,
  before skip=8pt, after skip=8pt,
  drop shadow={black!15},
}

\newtcblisting{toolpanel}[1]{%
  listing only,
  listing options={basicstyle=\ttfamily\scriptsize,breaklines=true,breakatwhitespace=true,columns=fullflexible,keepspaces=true,showstringspaces=false,frame=none,xleftmargin=0pt,xrightmargin=0pt,framexleftmargin=0pt,framexrightmargin=0pt,aboveskip=0pt,belowskip=0pt,literate={£}{{\pounds}}1},
  enhanced,
  arc=3pt, outer arc=3pt, boxrule=0.5pt,
  colback=toolbg, colframe=toolframe,
  colbacktitle=tooltitle, coltitle=codetitletext,
  fonttitle=\small\bfseries\sffamily,
  title={#1}, titlerule=0pt,
  toptitle=4pt, bottomtitle=4pt,
  top=5pt, bottom=5pt, left=6pt, right=6pt,
  boxsep=0pt,
  before skip=8pt, after skip=8pt,
  drop shadow={black!15},
}

\newpage

\section{Per-model Narratives}
\label{app:per-model}

 Here we provide detailed seed-level narratives for the five models evaluated across five seeds each in the betting benchmark. These
 supplement the failure-mode analysis in the main text with model-specific context on strategy, tool usage, and trajectory.

 \begin{table}[t]
 \centering
 \scriptsize
 \renewcommand{\arraystretch}{1.35}
 \setlength{\tabcolsep}{4pt}
 \begin{tabular}{@{}p{2.4cm}>{\raggedright\arraybackslash}p{3.0cm}>{\raggedright\arraybackslash}p{2.4cm}>{\raggedright\arraybackslash}p{1.5cm}>{\raggedright\arraybackslash}p{3.8cm}@{}}
 \toprule
 \textbf{Model} & \textbf{ML Approach} & \textbf{Staking (executed)} & \textbf{Adapt.} & \textbf{Characteristic Failure} \\
 \midrule
 GPT-5.4             & Walk-forward LR / RF+LR ensembles; one seed meta-rule bandit; Elo + rolling form & Rule-based flat \% (4--8\%); one
 seed pivoted to penny bets & Full & Front-loaded modelling; one seed conceded edge, reverted to capital preservation \\
 Claude Opus 4.6     & Calibrated GBM / RF+LR ensembles (41--63 features); one seed Poisson + market blend & Frac.\ Kelly 0.15--0.25
 (\pounds2--5 floor dominated) & Partial & Odds-as-features circularity diagnosed but never fixed; diagnostic insight without action \\
 GLM-5               & Bradley--Terry + Poisson; LogReg; RF+GB ensemble (varied by seed) & Fixed frac.\ 5\% / quarter-Kelly / full Kelly &
 Partial & Self-critique documents never implemented; context-window collapse in one seed \\
 Gemini 3.1 Pro      & RF on odds (base); XGBoost daily retrain (best); Elo+LogReg; LightGBM+EMA & Flat or frac-Kelly 0.05--0.10; one seed
 flat \pounds10 & Minimal & 3/5 bankrupt; base locked strategy on matchday~2; one seed accepted negative EV \\
 Kimi K2.5           & GBM with form features; inline Poisson xG; RF+GBM+LR ensemble & Kelly coded but rarely invoked; ad hoc flat in
 practice & None & Tool-call corruption; 98\%-of-bankroll accidental bet; 39 premature ``episode complete'' declarations \\
 \bottomrule
 \end{tabular}
 \caption{\textbf{Per-Model Strategy Summary.} Aggregated across five seeds per model. ``Staking'' reports what agents actually executed,
 not what they discussed in reasoning.}
 \label{tab:model-strategies}
 \end{table}

 \subsection{GPT-5.4 ($-7.9\%$ ROI, \pounds92{,}063).}
 The best performer on average and the most engineering-intensive model. All five seeds demonstrated strategy evolution, and three
 incorporated live model retraining using data downloaded by the environment after each matchday. No
  seed went bankrupt, though variance was large ($-32.9\%$ to $+34.1\%$). The best seed ($+34.1\%$ ROI) used a Random Forest and
 logistic regression with aggressive $8\%$ flat stakes, peaked at \pounds496 ($2.3\times$ starting bankroll), survived a ten-bet losing
 streak from March to April, and was rescued by a  \pounds8.64 bet on Arsenal versus Aston Villa at 10.0  that returned
 \pounds86.40. 
 
 \noindent The first seed devoted roughly 160 tool calls to pipeline construction before placing its first bet, building walk-forward
 validated logistic regressions with Elo ratings and rolling form features; after determining its model could not outperform the closing
 line (model log-loss 0.974 vs.\ market 0.971), it retreated to \pounds0.01 penny bets for $60\%$ of matchdays, preserving capital at the
 cost of any upside. The most strategically sophisticated seed ($-32.9\%$)
 implemented a meta-rule bandit tracking cumulative returns across four candidate betting rules and dynamically activating the best-scoring
 one, but finished among the worst GPT seeds. 
 
 \noindent Across seeds, GPT-5.4 discussed Kelly staking extensively in reasoning chains but uniformly
  implemented simpler percentage-of-bankroll rules in practice. For promoted teams, the first seed engineered a \texttt{home\_recent\_eng2}
 feature measuring the fraction of a team's last twenty matches in the Championship, effectively deferring to the market on teams it could
 not model.

 \subsection{Claude Opus 4.6 ($-11.2\%$ ROI, \pounds88{,}771).}
 The only model to avoid ruin across all five seeds, and the most consistent performer (range $-44.7\%$ to $+21.5\%$). Seeds consistently
 deployed calibrated gradient-boosting or random-forest ensembles with fractional Kelly staking. The best seed ($+21.5\%$) ran a calibrated RF+LR ensemble on autopilot
  with quarter-Kelly and an $8\%$ bankroll cap; its backtest reported $9{,}858\%$ ROI due to data leakage via
 \texttt{CalibratedClassifierCV} on the full training set, yet conservative sizing delivered genuine profit despite the absence of any
 mid-season adaptation. 
 
 \noindent The near-breakeven seed ($-0.3\%$) displayed the most impressive self-correction in the benchmark: after losing
 \pounds28 on Day~1, it diagnosed its GBM as dominated by home-team base rates, rebuilt the model as a Poisson-plus-market-prior blend with
 shrinkage toward bookmaker odds, and staged a V-shaped recovery from \pounds135 to \pounds219. Another seed ($-13.8\%$)
 discovered that draws at EV$>$1.20 yielded $15.4\%$ ROI in backtesting but placed only ten draw bets across the entire season. Across
 seeds, Opus showed a recurring tension around promoted teams: multiple seeds correctly identified model errors on Luton, Burnley, and
 Sheffield United fixtures but deferred to their models, reasoning that following the model-based framework was the right response.
 Two of five seeds showed meaningful mid-season adaptation; the other three ran identical pipelines from setup to season end, with the most
 profitable among them succeeding through conservative sizing rather than strategic insight.

 \subsection{GLM-5 ($-51.6\%$ ROI, \pounds48{,}395).}
 GLM-5 displayed a distinctive and recurring pattern across all five seeds: accurate self-diagnosis followed by no corrective action. The
 best seed ($-14.3\%$) used a hand-built Bradley--Terry model with Poisson goal predictions, recovering from a trough of \pounds93 by
 tightening edge thresholds, and wrote three self-critique documents correctly identifying its hardcoded $25\%$ draw rate, overestimation of
  home advantage, and miscalibration on promoted teams. Notably it translated none of these diagnoses into model corrections. One seed ($-29.7\%$)
  was the partial exception: it retrained its Random Forest each matchday and peaked at \pounds464 from longshot wins, but gave back all
 gains chasing underdogs in the final fifteen matchdays, collapsing to \pounds155. 
 
 \noindent The bankrupt seed ($-100\%$) used a Random Forest plus
 gradient boosting ensemble with quarter-Kelly staking, surviving to matchday~76 until bankruptcy. Another seed ($-62.1\%$) trained a logistic regression with $50.5\%$
 accuracy, acknowledged on Day~1 that this was barely above baseline, hardcoded a skip-Luton rule while continuing to lose on Burnley and
 Sheffield United, and declared its $-55\%$ drawdown a ``successful completion''. Across seeds, GLM-5 generated the most extensive written
 self-analysis of any model, including multiple markdown reports per seed documenting calibration failures and proposed fixes yet showed the widest
  gap between metacognitive ability and actual corrective behaviour.

 \subsection{Gemini 3.1 Pro ($-66.0\%$ ROI, \pounds34{,}029).}
 This model exhibited the widest seed variance: three of five seeds went bankrupt while one produced the single best seed-level ROI in the
 benchmark ($+33.7\%$). The profitable seed used two XGBoost classifiers with daily retraining and disciplined $5\%$ fractional Kelly
 sizing (the only Gemini seed to retrain its model). A critical flaw in its staking code (calculating Kelly fractions against a hardcoded
 \pounds200 pseudo-bankroll rather than the live bankroll) prevented compounding. The base seed ($-63.6\%$) trained a Random Forest solely on five bookmaker-odds columns, producing predictions that added noise
 rather than signal; after a fortunate early \pounds100 flat bet, it locked this strategy on matchday~2 and ran on autopilot for the remainder of the season. Among bankrupt seeds, one used an Elo plus logistic regression with no EV
 threshold, placing thirteen simultaneous wagers on matchday~2 totalling $38\%$ of
 its bankroll and entering an unrecoverable decline. Another deliberately relaxed its EV threshold to $-10\%$ then entered a $3{,}500$-line infinite loop cycling
 \texttt{view\_matches}/\texttt{view\_bankroll} post-bankruptcy. The third bankrupt seed used a static nine-feature Random Forest with flat
 \pounds10 stakes, wrote Kelly code it never executed, and rationalised its bankruptcy as having ``successfully met the system's
 objective''. Gemini's results demonstrate the importance of retraining: the single seed that incorporated new data achieved
 $+33.7\%$; the four that did not averaged $-91\%$.

 \subsection{Kimi K2.5 ($-89.6\%$ ROI, \pounds10{,}421).}
 The model whose failures were primarily infrastructural rather than strategic. No seed finished profitably, and two of five went bankrupt.
 One seed ($-77.7\%$) used a GBM with quarter-Kelly that ran on extended autopilot; it declared the episode ``complete'' six times
 mid-season, entering an eighteen-matchday zombie plateau with no bets placed, before resuming with placeholder \pounds0.001 wagers. Another
  seed achieved a $62.3\%$ win rate - the highest of any seed in the benchmark - using an inline Poisson expected-goals model that peaked at
  \pounds374 ($+70\%$), but abandoned its model from mid-December and suffered tool-call corruption
 that terminated the active session; its bankroll bled through forced minimum bets to \pounds25 by season end. A third seed froze a GBM on
 $1993$--$2023$ cumulative statistics, predicting a $233\%$ edge on bottom-of-league Sheffield United based on the club's all-era historical
  win rate; it declared the episode ``complete'' thirty-nine separate times while its bankroll fell from \pounds220 to \pounds40. The
 bankrupt first seed built a GBM pipeline with eighteen features but never executed its prediction script; approximately fifty consecutive
 tool calls produced garbage commands (\texttt{": ["}), and a single malformed call placed \pounds251 ($98\%$ of the bankroll) on Burnley
 versus Luton. The final seed removed its probability threshold early, allowing longshot bets with sub-$30\%$ win probability, then overrode
  its own Kelly calculation to go all-in on its final bet. Across seeds,
 Kimi's Kelly implementations were textbook-correct but the probability inputs from miscalibrated or never-executed models rendered them
 ineffective.

\section{Example Matchday Loop: Claude Opus 4.6, Matchday 1}
\label{app:example-loop}

To give a concrete sense of the agent--environment interaction, we reproduce the full tool-call loop for the opening matchday of Claude Opus 4.6's best seed ($-0.2\%$ ROI). The loop has five stages: observe fixtures, run the trained model, reason about value, place bets, and settle. Earlier in the episode the agent had already trained a gradient-boosting and Random Forest ensemble over ten seasons of historical data (\texttt{model.py}) and written a betting-strategy script (\texttt{bet\_today.py}) that applies the model to the current matchday and reports any value bets found. The loop below begins with the single match on 11/08/2023.

\begin{toolpanel}{\faWrench~~\texttt{view\_matches()}}
Matches for 11/08/2023:

Match 0: Burnley vs Man City
  Odds - Home: 8.0, Draw: 5.5, Away: 1.33
  Over 2.5: 1.67, Under 2.5: 2.2
\end{toolpanel}

\begin{toolpanel}{\faWrench~~\texttt{bash(command="python3 bet\_today.py")}}
Current Bankroll: £220.00
Strategy: ML Value Betting with Quarter-Kelly Sizing

Analyzing: Burnley vs Man City (Match 0)

Model Predictions:
  Home win: 40.1% (odds: 8.0, implied: 12.5%)
  Draw:     26.9% (odds: 5.5, implied: 18.2%)
  Away win: 33.0% (odds: 1.33, implied: 75.2%)

  VALUE BETS FOUND:
    HOME: edge=220.7%, model_prob=40.1%, kelly=7.9%, bet=£17
    DRAW: edge=48.2%, model_prob=26.9%, kelly=2.7%, bet=£6
    UNDER_2_5: edge=8.9%, model_prob=49.5%, kelly=1.9%, bet=£5

  Total exposure: £28 (12.7% of bankroll)
\end{toolpanel}

\begin{reasonpanel}{\faComment[regular]~~Assistant reasoning}
The model has identified value bets. Now let me place them based on the ML model's recommendations.
\end{reasonpanel}

\begin{toolpanel}{\faWrench~~\texttt{place\_bet(match\_id=0, bet\_type="home", amount=17)}}
Bet placed: 17.0 on home for Burnley vs Man City at odds 8.0.
Potential return: 136.00
\end{toolpanel}

\begin{toolpanel}{\faWrench~~\texttt{place\_bet(match\_id=0, bet\_type="draw", amount=6)}}
Bet placed: 6.0 on draw for Burnley vs Man City at odds 5.5.
Potential return: 33.00
\end{toolpanel}

\begin{toolpanel}{\faWrench~~\texttt{place\_bet(match\_id=0, bet\_type="under\_2\_5", amount=5)}}
Bet placed: 5.0 on under\_2\_5 for Burnley vs Man City at odds 2.2.
Potential return: 11.00
\end{toolpanel}

\begin{toolpanel}{\faWrench~~\texttt{next\_matchday()}}
Bet Results:

x LOST: Burnley vs Man City (home) - Lost: £17.00
x LOST: Burnley vs Man City (draw) - Lost: £6.00
x LOST: Burnley vs Man City (under_2_5) - Lost: £5.00

Net result: £-28.00
New bankroll: £192.00

Advanced to next matchday: 12/08/2023
\end{toolpanel}

\vspace{1em}

\noindent The opening matchday illustrates a characteristic early-season failure: the model assigned $40.1\%$ to a Burnley home win against an implied $12.5\%$, claiming a $220\%$ edge on a newly promoted side facing the reigning champions.

\newpage
\section{Pitfalls of Non-stationarity}

\begin{figure}[H]
\centering
    \includegraphics[width=1.00\linewidth]{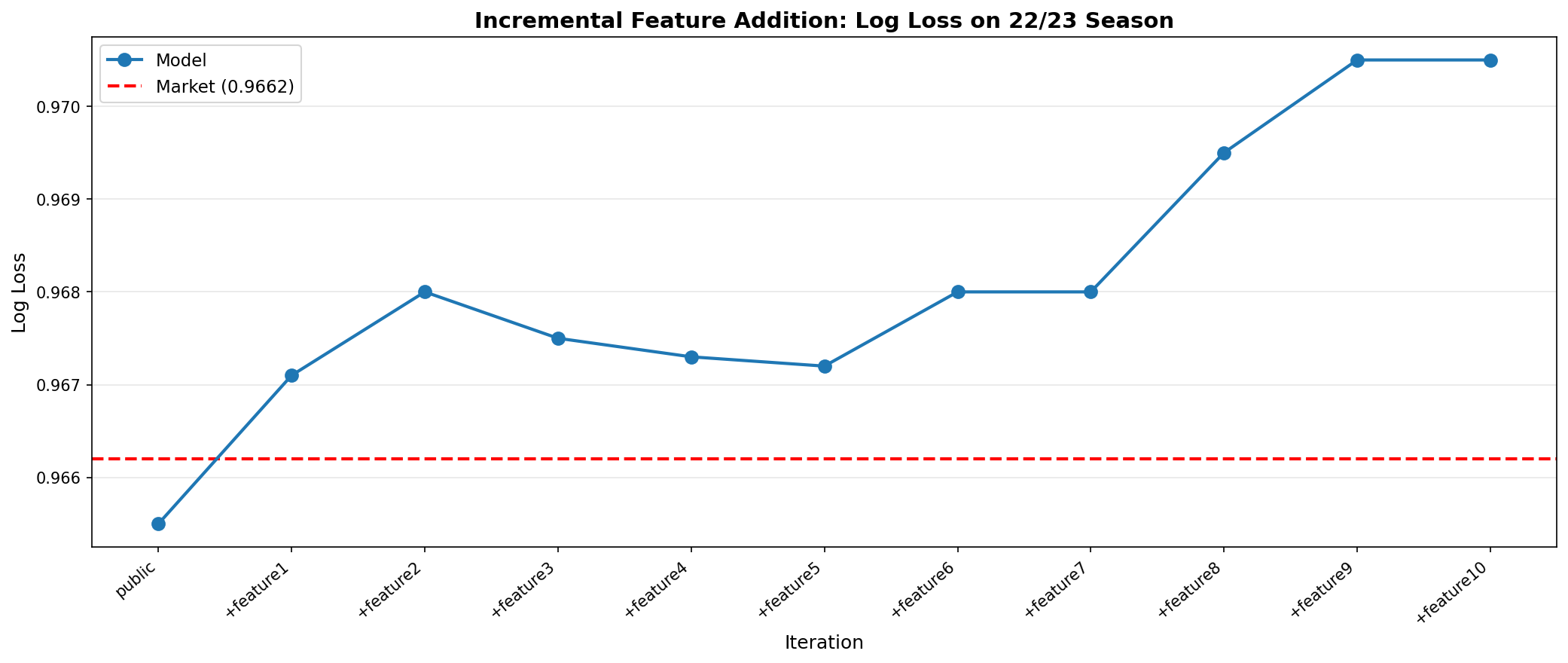}
    \caption{\textbf{Feature Instability in Non-stationary Environments}. Models typically split their historical data into training and validation sets, but do little work on features stability analysis. This problem becomes more pronounced with recent data in the 19/20 and 20/21 seasons which had Covid-era effects, including crowd restrictions and more. Above we see out-of-sample performance from a feature selection procedure, in which every feature the model added based on validation set performance failed to generalise out-of-sample - showing the difficulty of the task compared to traditional machine learning competitions without non-stationarity.}
\label{fig:sigmoid_steepness_comparison}
\end{figure}

\newpage
\section{3 Seed Benchmark}

\begin{figure}[H]
\centering
    \includegraphics[width=1.00\linewidth]{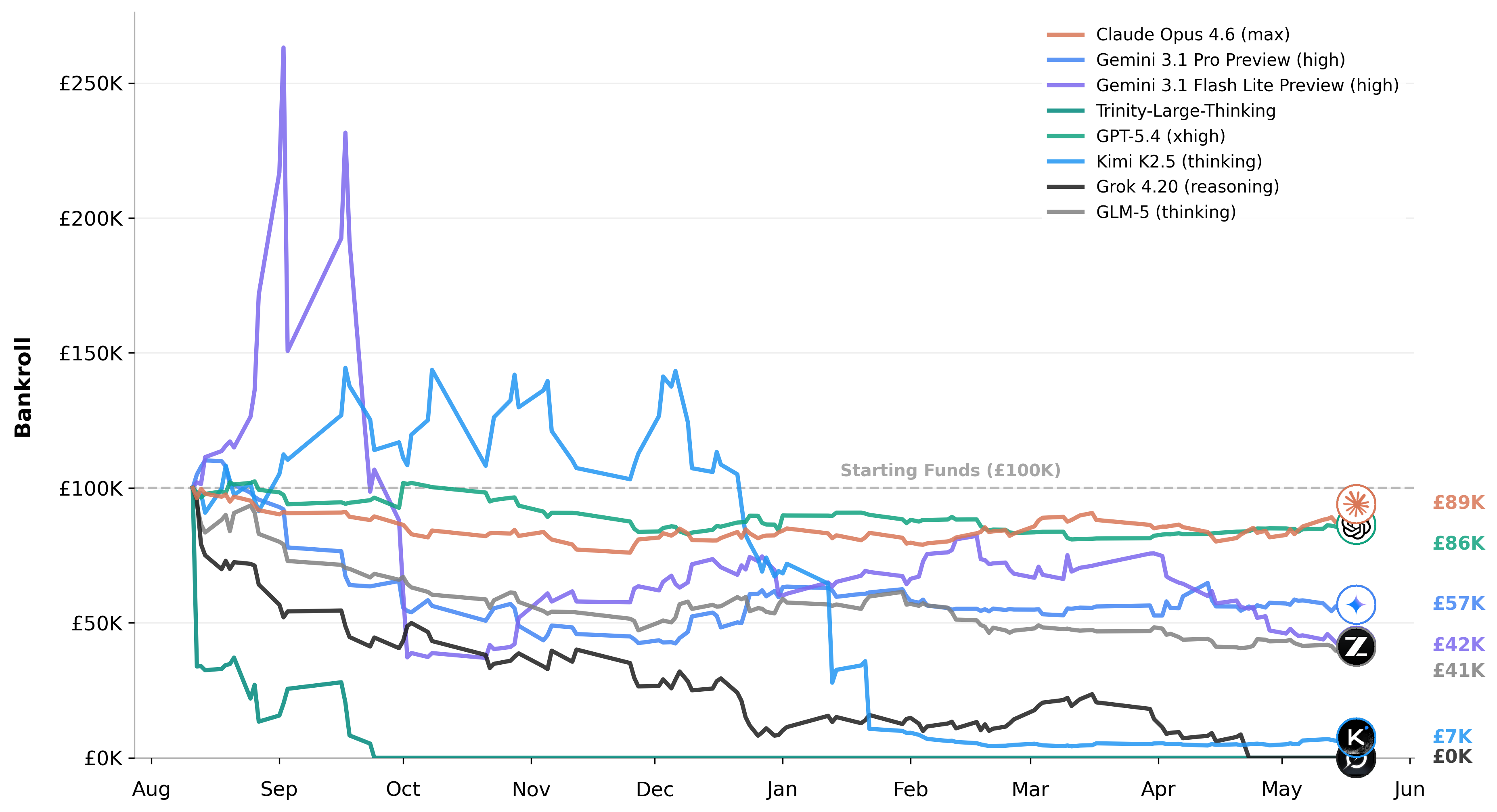}
    \caption{\textbf{Model Performance on KellyBench}. KellyBench tasks models with developing machine learning betting strategies for the 2023/24 English Premier League season with the goal of maximising long-term bankroll growth. No model makes a return on average across 3 seeds. Models also fail to adapt strategies in response to failure. Initial bankroll is normalised to £100K for display purposes.}
\label{fig:sigmoid_steepness_comparison}
\end{figure}

The initial version of \texttt{KellyBench} (v1) used three seeds. We also evaluated Grok 4.20 and Arcee but found that these models struggled to complete the season. For v1.1 we increased the number of seeds to 5 to increase pairwise statistical significance between models, and also excluded models that struggled to complete the season.

\newpage
\section{Tool Specifications}\label{app:tools}

Here we document the tools available to agents in the betting environment. Agents interact with a simulated Premier League season by viewing fixtures, placing bets, managing bankroll, and advancing through matchdays. In addition, a CLI toolset mirroring the Claude Code interface is available for model development within a sandboxed compute environment.

\subsection{\texttt{view\_matches}}
\noindent\textbf{Purpose.} Display the fixtures and bookmaker odds for the current matchday.

\noindent\textbf{Parameters.} None.

\noindent\textbf{Returns.} A formatted list of matches for the current date. Each entry includes the match ID, home and away team names, and decimal odds for the home/draw/away markets (sourced from Bet365, Gamebookers, or Interwetten in fallback order). Where available, over/under 2.5 goals odds are also displayed.

\subsection{\texttt{place\_bet}}
\noindent\textbf{Purpose.} Place a wager on a specific match outcome.

\noindent\textbf{Parameters.}
\begin{itemize}[nosep,leftmargin=1.5em]
    \item \texttt{match\_id} (int): Index of the match on the current matchday, as returned by \texttt{view\_matches}.
    \item \texttt{bet\_type} (str): One of \texttt{"home"}, \texttt{"draw"}, \texttt{"away"}, \texttt{"over\_2\_5"}, or \texttt{"under\_2\_5"}.
    \item \texttt{amount} (float): Stake in pounds. Must be positive and at most the current bankroll.
\end{itemize}

\noindent\textbf{Returns.} Confirmation of the placed bet, including the matched odds and potential return. The stake is immediately deducted from the bankroll. Multiple bets may be placed per matchday (including on the same match).

\subsection{\texttt{view\_bankroll}}
\noindent\textbf{Purpose.} Query the agent's current financial state.

\noindent\textbf{Parameters.} None.

\noindent\textbf{Returns.} The current bankroll balance. If bets have been placed on the current matchday, also reports the total value staked and the amount still available for betting.

\subsection{\texttt{next\_matchday}}
\noindent\textbf{Purpose.} Settle all pending bets and advance the simulation to the next set of fixtures.

\noindent\textbf{Parameters.} None.

\noindent\textbf{Precondition.} At least one bet must have been placed on the current matchday; the tool refuses to advance otherwise.

\noindent\textbf{Returns.} A settlement report listing each bet as won or lost, with stakes and payouts. Winning bets pay out at the quoted decimal odds (stake $\times$ odds). The report includes the net result, the updated bankroll, and the date of the next matchday. Updated match data (results, player statistics) is downloaded to \texttt{/home/ubuntu/latest\_data/} for use in model retraining. When the final matchday is settled, the environment terminates and reports the final bankroll and total profit/loss.

\noindent\textbf{Reward signal.} The per-step reward is the log wealth ratio $\ln(B_{t+1} / B_t)$, where $B_t$ is the bankroll before settlement and $B_{t+1}$ is the bankroll after. This incentivises Kelly-optimal growth.

\subsection{\texttt{bash}}
\noindent\textbf{Purpose.} Execute arbitrary shell commands in a sandboxed Linux environment for data analysis and model development.

\noindent\textbf{Parameters.}
\begin{itemize}[nosep,leftmargin=1.5em]
    \item \texttt{command} (str): The shell command to execute.
    \item \texttt{description} (str, optional): A brief note on the command's intent.
    \item \texttt{timeout} (float, optional): Maximum execution time in seconds (default 30).
\end{itemize}

\noindent\textbf{Returns.} Standard output and the exit code. Network access is blocked: commands matching common HTTP/socket patterns (e.g.\ \texttt{curl}, \texttt{requests}, \texttt{urllib}) are rejected, and three such attempts result in disqualification with bankroll elimination. Historical match data is pre-loaded at \texttt{/tmp/gr-datasets/}.

\subsection{\texttt{read}}
\noindent\textbf{Purpose.} Read the contents of a file in the sandbox.

\noindent\textbf{Parameters.}
\begin{itemize}[nosep,leftmargin=1.5em]
    \item \texttt{file\_path} (str): Path to the file to read.
    \item \texttt{offset} (int, optional): Starting line number.
    \item \texttt{limit} (int, optional): Maximum number of lines to return.
\end{itemize}

\noindent\textbf{Returns.} Numbered file contents (or the requested slice). When both \texttt{offset} and \texttt{limit} are provided, returns lines in the range $[\texttt{offset},\; \texttt{offset} + \texttt{limit}]$.

\subsection{\texttt{write}}
\noindent\textbf{Purpose.} Create or overwrite a file with the given content.

\noindent\textbf{Parameters.}
\begin{itemize}[nosep,leftmargin=1.5em]
    \item \texttt{file\_path} (str): Destination path. Parent directories are created automatically.
    \item \texttt{content} (str): The full file content to write.
\end{itemize}

\noindent\textbf{Returns.} Confirmation of successful write.

\subsection{\texttt{edit}}
\noindent\textbf{Purpose.} Perform exact string replacement within an existing file.

\noindent\textbf{Parameters.}
\begin{itemize}[nosep,leftmargin=1.5em]
    \item \texttt{file\_path} (str): Path to the file to edit.
    \item \texttt{old\_string} (str): The substring to find. Must appear exactly once unless \texttt{replace\_all} is set.
    \item \texttt{new\_string} (str): The replacement string.
    \item \texttt{replace\_all} (bool, optional): If true, replaces all occurrences. Default false.
\end{itemize}

\noindent\textbf{Returns.} Confirmation of successful edit. Fails if \texttt{old\_string} is not found or is ambiguous (appears more than once with \texttt{replace\_all=false}).

\subsection{\texttt{glob}}
\noindent\textbf{Purpose.} Find files matching a glob pattern.

\noindent\textbf{Parameters.}
\begin{itemize}[nosep,leftmargin=1.5em]
    \item \texttt{pattern} (str): Filename glob pattern (e.g.\ \texttt{"*.csv"}).
    \item \texttt{path} (str, optional): Root directory to search from. Defaults to the working directory.
\end{itemize}

\noindent\textbf{Returns.} A sorted list of matching file paths.

\subsection{\texttt{grep}}
\noindent\textbf{Purpose.} Search for a regex pattern across files.

\noindent\textbf{Parameters.}
\begin{itemize}[nosep,leftmargin=1.5em]
    \item \texttt{pattern} (str): Regular expression to match.
    \item \texttt{path} (str, optional): Root directory. Defaults to the working directory.
    \item \texttt{glob} (str, optional): If provided, restricts the search to files matching this glob pattern.
\end{itemize}

\noindent\textbf{Returns.} Matching lines with file paths and line numbers.

\subsection{\texttt{todo\_write}}
\noindent\textbf{Purpose.} Manage a structured to-do list for task planning and progress tracking.

\noindent\textbf{Parameters.}
\begin{itemize}[nosep,leftmargin=1.5em]
    \item \texttt{todos} (list[dict]): A list of to-do items, each with fields \texttt{id}, \texttt{content} (str), \texttt{status} (one of \texttt{"pending"}, \texttt{"in\_progress"}, \texttt{"completed"}), and \texttt{priority} (one of \texttt{"high"}, \texttt{"medium"}, \texttt{"low"}).
\end{itemize}

\noindent\textbf{Returns.} A formatted rendering of the current to-do list. The list is replaced in full on each call (i.e.\ last-write-wins).